\documentclass[final]{clv2025}

\jvol{vv}
\jnum{nn}
\jyear{2025}
\usepackage{amsmath}
\usepackage{booktabs}

\usepackage{graphicx} % For including graphics
\usepackage{caption}  % For customising captions
\usepackage{float}    % For better figure control
\usepackage{tcolorbox}
\usepackage{hyperref}
\usepackage{todonotes}
\usepackage{color, colortbl}
\definecolor{Gray}{gray}{0.9}
\usepackage{multirow,adjustbox}

\usepackage{diagbox}

\newcolumntype{M}[1]{>{\centering\arraybackslash}m{#1}}
\usepackage{colortbl}
\usepackage{xcolor}
\usepackage{sidecap}

\runningtitle{}
\runningauthor{}

\begin{document}

\title{Multilingual LLMs Struggle to Link Orthography and Semantics in Bilingual Word Processing}

\author{Eshaan Tanwar\thanks{Equal contribution}$^{,1}$, Gayatri Oke$^{*,1}$, Tanmoy Chakraborty\thanks{Corresponding authors: tanchak@iitd.ac.in}$^{,1,2}$}

\affilblock{
    \affil{Department of Electrical Engineering, Indian Institute of Technology Delhi, New Delhi, India\\}
    \affil{Yardi School of Artificial Intelligence, Indian Institute of Technology Delhi, New Delhi, India}
}

\maketitle

\begin{abstract}
Bilingual lexical processing is shaped by the complex interplay of phonological, orthographic, and semantic features of two languages within an integrated mental lexicon. In humans, this is evident in the ease with which cognate words -- words sharing both orthographic form
and meaning (e.g., \textit{blind}, meaning "sightless'' in both English and German) -- are processed, compared to the challenges posed by interlingual homographs, which share orthographic form but differ in meaning (e.g., \textit{gift}, meaning "present'' in English but "poison'' in German). We investigate how multilingual Large Language Models (LLMs) handle such phenomena, focusing on English-Spanish, English-French, and English-German cognates, non-cognates, and interlingual homographs. Specifically, we evaluate their ability to disambiguate meanings and make semantic judgments, both when these word types are presented in isolation and within sentence contexts. Our findings reveal that while certain LLMs demonstrate strong performance in recognising cognates and non-cognates in isolation, they exhibit significant difficulty in disambiguating interlingual homographs, often performing below random baselines. This suggests that LLMs tend to rely heavily on orthographic similarities rather than semantic understanding when interpreting interlingual homographs. Further, we find that LLMs exhibit difficulty in retrieving word meanings, with performance in isolative disambiguation tasks having no correlation with semantic understanding. Finally, we study how LLMs process interlingual homographs in incongruent sentences. We find models to opt for different strategies in understanding English and non-English homographs, highlighting a lack of a unified approach to handling cross-lingual ambiguities.
\end{abstract}

\section{Introduction}

The advent of Large Language Models (LLMs) has enhanced the ability of artificial systems to model natural language. These sophisticated models have nearly achieved human-like performance in grammar and sentence comprehension~\cite{cai2023does}. However, studies evaluating the linguistic and cognitive capabilities of these models focus primarily on English~\cite{lai2023chatgpt,kosinski2023theory, zhang2023don}. Given that almost half of the global population speaks at least two languages~\cite{grosjean2021life}, it is pertinent to assess the bilingual capabilities of these models and whether their processing proficiency is consistent across languages. This would also imply their ability in multilingual processing and the challenges faced there, which is a crucial, yet relatively undetermined area of study.

\begin{figure*}%[tbhp]
\centering
\includegraphics[width=\linewidth]{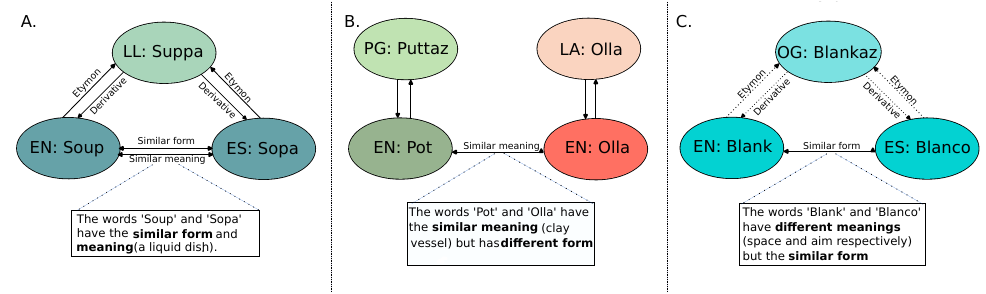}
\caption{{\bf A working example to illustrate three linguistic entities: cognates, non-cognates and interlingual homographs}, using words from Late Latin (LA), English (EN), Spanish (ES), Proto-Germanic (PG), Latin (LL), and Old German (OG). \textbf{(A)} \textbf{Cognates:} The English word `Soup' and the Spanish word `Sopa' share the same meaning (a liquid dish) and similar orthography, derived from the Latin root `Suppa'. \textbf{(B)} \textbf{Non-cognates}: The English word `Pot' and the Latin word `Olla' represent the same meaning (a clay vessel) but have different orthography, tracing their origins to distinct roots (Proto-Germanic: `Puttaz' and Late Latin: `Olla'). \textbf{(C)} \textbf{Interlingual Homographs:} The English word `Blank' and the Spanish word `Blanco' differ in meaning (space/gap vs. goal/aim) but \textcolor{black}{display similar orthography}, derived from the Old German root `Blankaz.' However, homographs don't need to share an etymological root; hence, they are marked by a dotted line in the figure.}
\label{fig:cog_example}
\end{figure*}

\textcolor{black}{As explained by \citet{traxler2011introduction}, existing research on human cognition indicates that lexical access is a key component of bilingual processing.} This refers to various processes and mental representations that are involved in identifying a word in a given language. The semantic information about the word's meaning in that language is activated only after it has been recognised. However, bilingual lexical access is influenced by whether a word belongs to the category of cognates, non-cognates, or interlingual homographs. Cognates are words with similar orthography and the same meaning in two languages (Figure~\ref{fig:cog_example}). In contrast, interlingual homographs are words that appear \textcolor{black}{similar} in form but have different meanings in two languages (Figure~\ref{fig:cog_example}). Non-cognates are words with the same meaning but different forms in two languages, \textcolor{black}{and are used in translations} (Figure~\ref{fig:cog_example}).  Psycholinguistic studies involving picture naming, \textcolor{black}{word} translation, and lexical decision tasks have explored how these word categories facilitate or hinder bilingual processing~\cite{caramazza1979lexical,costa2000cognate}. In a lexical decision task (LDT), for example, participants are asked to determine whether the target word (which may be a cognate or an interlingual homograph) \textcolor{black}{belongs to, and is meaningful in} the language in which they are instructed to respond~\cite{jared2001bilinguals}. \textcolor{black}{Depending on whether the target words are presented in isolation or used in sentences, insight can be gained into the impact of word type on bilingual word recognition and processing~\cite{dijkstra2015sentence}. In our study, we devise three novel tasks aimed at understanding LLMs' bilingual word-processing ability by utilising cognates, non-cognates and interlingual homographs. We examine the processing abilities of five publicly-available multilingual LLMs by testing English-Spanish, English-French, and English-German target word pairs, both in isolation and within sentence contexts. Specifically, we evaluate whether these models can store and retrieve the meaning of target words and disambiguate between the meanings of interlingual homographs when provided with a semantically constrained sentence context. Analysing how multilingual LLMs process cognates, non-cognates, and interlingual homographs can provide insight into bilingual lexical access as a computational process. Understanding these patterns in LLMs can thus inform the development of bilingual interactive AI systems, allowing users to communicate naturally in multiple languages without compromising comprehension or response quality. Such findings could guide model design, prompting improvements in cross-lingual understanding, context-sensitive word disambiguation, and generation in multilingual settings.}

\paragraph{\bf Bilingual lexical access and processing}

Psycholinguistic theories propose that information regarding different components of words, such as their phonology, orthography, morphology, and meaning, is captured and stored in the mental lexicon~\cite{traxler2011introduction}. For bilingual speakers, their knowledge of two languages is stored in the lexicon without any partition. This suggests that the bilingual lexicon is integrated, and that processing doesn't occur via separate, independent routes for the two languages~\cite{ grosjean1989neurolinguists}. Instead, the integrated lexicon is accessed whenever necessary through a non-selective process that makes the knowledge of both languages available at the same time~\cite{dijkstra2002architecture}. For instance, when presented with an object that activates a particular meaning or conceptual link, bilingual speakers have access to words from both languages that represent this meaning (Figure~\ref{fig:cog_example}). The selection of a word ultimately depends on its level of activation, which is influenced by linguistic context and external factors such as task demands \textcolor{black}{or language proficiency.}

Different types of words exert their unique influence on bilingual processing by making lexical representations from both languages compete with each other for greater activation in the integrated lexicon. In the case of cognates, this framework facilitates word processing~\cite{dijkstra2002architecture}. \textcolor{black}{Cognate words are derived from a common etymological root, which results in them sharing similar phonological features and orthography (Figure~\ref{fig:cog_example}). In addition to this surface-level commonality, cognates share the same meaning and thus represent a common semantic or conceptual link. In the integrated bilingual lexicon, this leads to stronger activation of the lexical representations of the cognate words, which makes them easier to retrieve during word processing.} This is known as the \emph{cognate facilitation effect}~\cite{costa2000cognate, sherkina2003cognate}, as explained in the Bilingual Interactive Activation Plus (BIA+) Model. Contrarily, interlingual homographs are deceptive and often impede processing. Since these words are similar only in terms of their form and hold distinct meanings across the two languages, they have separate, yet partially overlapping representations in the integrated lexicon. According to the BIA+ model, when a homograph is encountered, both representations and their corresponding meaning links are simultaneously activated and compete for selection. This competition escalates with increasing similarity between the orthographic-phonological forms of the two words, further complicating the task of assigning the correct meaning to each of them. Consequently, it is more difficult and time-consuming to process interlingual homographs, especially when they are presented in isolation or with minimal contextual cues. 

\paragraph{\bf Multilingual large language models} 

\textcolor{black}{With approximately 7,000 languages spoken worldwide, linguistic diversity presents a significant challenge in developing intelligent systems capable of processing and responding in multiple languages~\cite{joshi-etal-2020-state}.} LLMs offer a promising solution to this challenge with their ability to generalise across languages, enabling robust performance even in low-resource settings. Their few-shot learning capabilities further enhance their adaptability, making them powerful tools for building cross-lingual systems~\cite{winata2021language,schick-schutze-2021-just,lin-etal-2022-shot,Steyvers2025}. However, how these models acquire their multilingual capabilities is not fully understood. LLMs are not trained on parallel corpora across languages; instead, their training data consists of a mixture of different languages in varying proportions and their performance is a by-product of unsupervised training. 

Existing research suggests that shared vocabulary of words that are common across languages plays a pivotal role in enhancing multilingual and cross-lingual performance in language models~\cite{chung-etal-2020-improving,conneau-etal-2020-emerging,zheng-etal-2021-allocating,yuan-etal-2024-vocabulary}. These shared lexical items serve as anchor points, aligning different languages within a unified representational space and facilitating cross-linguistic transfer. Cognates, due to their \textcolor{black}{similar} form and meaning across languages, strengthen this alignment by reinforcing structural and semantic parallels. In contrast, interlingual homographs, with their misleading surface similarity and distinct meanings across languages, pose a crucial challenge for semantic disambiguation. This is akin to the cognate facilitation effect observed in humans, alluding to cognates sharing a common representation space that makes word retrieval faster. Interlingual homographs introduce complexity in word recognition by interfering with semantic processing, similar to the inhibition effect they cause in human cognition. 

% Analyzing how models process these elements offers valuable insights into their underlying generalisation mechanisms and cross-lingual capabilities, advancing our understanding of multilingual natural language processing. 

In this work, we analyse these linguistic features in LLMs to better understand their bilingual word-processing capabilities. Our approach draws inspiration from psycholinguistic experiments, particularly lexical decision tasks, which are widely employed to study multilingual processing in humans. These tasks utilise audio-visual cues to measure response times and accuracy in tasks such as translation, sentence comprehension, and word recognition, focusing on cognates, non-cognates, and interlingual homographs. However, since LLMs process only textual data, these tasks cannot be directly applied. We adapt them to evaluate the cross-lingual and multilingual capabilities of LLMs and draw parallels with human cognition \cite{Mischler2024,Tikochinski2025}. 

\paragraph{\bf Reproducibility} Our dataset, which includes words along with their meanings and semantic constraint sentences for cognates, non-cognates, and interlingual homographs across the three language pairs (English-Spanish, English-French, and English-German), has been released for further utility. In addition, the source code for the analyses conducted in this paper is available at \href{https://github.com/EshaanT/Bilingual_processing_LLMs}{https://github.com/EshaanT/Bilingual\_processing\_LLMs}.

% We examine LLMs multilingual ability via  the following research questions (RQs):

% \begin{itemise}
%     \item \textbf{RQ1:} How proficient are multilingual LLM's in disambiguating cognates, non-cognates, and interlingual homographs?
%     \item \textbf{RQ2:} Can multilingual LLM's correctly store the meaning of cognates, non-cognates and interlingual homographs?
%     \item \textbf{RQ3:} How do multilingual LLM's deal with code-mixed sentences through semantic constraints?
% \end{itemise}

\section{Related Work}
\textcolor{black}{Psycholinguistic research over the last few decades has primarily focused on understanding lexical access in speech production. Bilingual lexical access, in particular, has attracted the attention of many researchers and motivated them to propose various theories and models of word processing in the integrated lexicon. Among these, the most widely accepted explanation of lexical access involves the principle of spreading activation. As explained by \citet{costa2000lexical}, it is hypothesised that when humans are presented with a stimulus that activates a certain semantic or conceptual link, there is simultaneous activation of concepts that are related to the original stimulus. This activation spreads from the semantic level to lexical-level representations, and finally, to the phonological features of individual words. These activated lexical nodes compete for selection, and only the node with the highest proportion of activation is retrieved and produced. In the case of bilingual speakers, the presented stimulus simultaneously activates semantic links from both languages, which then follow the same pattern of activation and competition for selection.}

\textcolor{black}{While the principle of spreading activation and lexical selection is widely agreed upon, there is a difference of opinion on the extent to which activation can spread before the process of lexical selection actually takes place. Discrete serial models of lexical access propose that once a semantic link is activated by a stimulus, its activation spreads to the lexical-level representations of words related to that concept. However, it does not extend to the phonological level until one of the activated lexical nodes is selected. Additionally, activation at the phonological level occurs only for the selected lexical node \cite{levelt1993speaking, levelt1991time}. In contrast, cascaded activation models of lexical access propose that activation flows from the semantic level to the phonological level, even for non-selected lexical nodes, and that lexical selection occurs only after the spread of activation \cite{caramazza1997many, dell1986spreading, jescheniak1998discrete}. Most importantly, these models assume that the process of lexical selection is not language-specific, meaning that for bilingual speakers, lexical nodes from both languages receive parallel activation and compete for selection \cite{dijkstra2002architecture, peterson1998lexical}.}

\textcolor{black}{Experimental evidence over the years has shown overwhelming support for the cascaded activation view of lexical access. \citet{costa2000lexical} explained that when a given semantic link (such as \textit{“dog”}) and other concepts related to it (such as \textit{“cat”}) are activated, the phonological features of both lexical nodes are simultaneously made available for the speaker to select. Usually, the speaker chooses the node with the highest level of activation to retrieve the target word (in this case, \textit{“dog”}). However, the competition between these activated nodes can also cause spontaneous speech errors due to the interference of the non-target node (in this case, \textit{“cat”}). Specifically for bilingual lexical access, cascaded activation models are able to explain how bilingual speakers retrieve words from their integrated lexicon.}

In particular, the Bilingual Interactive Activation Plus (BIA+) model \cite{dijkstra2002architecture} highlights the special status of cognate words by explaining how cascaded activation influences bilingual processing. When multiple lexical nodes across two languages and corresponding to the same semantic representation are activated simultaneously, the orthographic-phonological overlap between cognate words (e.g., \textit{"yogurt"} in English and \textit{"yogur"} in Spanish) is expected to increase overall activation, even of the non-selected lexical node \cite{costa2000cognate}. This makes cognate words much easier for bilingual speakers to select, ultimately resulting in faster retrieval and articulation.

\textcolor{black}{\citet{costa2000cognate} experimentally tested this with the help of a picture-naming task that was administered to bilingual Spanish-Catalan speakers, as well as monolingual Spanish speakers.} All participants were presented with pictures that had either cognate or non-cognate names in the two languages and were asked to name them as quickly and as accurately as possible. Upon comparing the naming latencies for both groups of participants, it was found that the cognate words were named much faster by bilingual speakers than the non-cognate words. In contrast, no such difference was observed in the performance of monolingual speakers. This was understood as the cognate facilitation effect, where the simultaneous activation of cognate lexical nodes resulted in greater retrieval and production efficiency. 

In addition to such simultaneous cross-linguistic activation, sentence context was found to influence bilingual word recognition and processing. This was proved with the help of another picture-naming task where bilingual Dutch-English speakers were asked to name cognate or non-cognate words in visually presented sentence contexts~\cite{starreveld2014parallel}. Similar to the phenomenon observed in out-of-context picture naming, the participants were able to name cognate words faster than non-cognate words, thereby supporting the view that cognate facilitation persists even within sentence contexts. Another study extended these findings by designing a lexical decision task that was administered to bilingual Dutch-English speakers~\cite{dijkstra2015sentence}. The participants were asked to read sentences in Dutch (their L1) or English (their L2). Each sentence was supposed to be read word-by-word and ended in a target word that the participants had to classify as a lexical item or not in the given language. The target words were cognates or non-cognates, and the sentences varied in their level of semantic constraint to manipulate how predictable the final word would be. Although cognate facilitation was modulated to some extent by native language influence and the level of semantic constraint, it could not be eliminated. Thus, cognate words were still found to be recognised faster than non-cognate words due to their simultaneous activation during the word recognition process.  

The co-activation of various lexical nodes is evidently advantageous to bilingual processing in the case of cognate words. However, with respect to interlingual homographs, this leads to significant interference and eventually inhibits efficient word processing. Picture-naming or word-translation experiments show bilingual speakers responding more slowly to interlingual homographs due to their deceptive orthographic and phonological similarities~\cite{dijkstra1999recognition}. Simultaneous activation of the semantic link of a homograph in the non-target language interferes with the language recognition process of the speakers, ultimately causing a delay in their response times~\cite{costa2000cognate, hermans1998producing}.  This delay persists even when the homographs are presented within a sentence context. 

\citet{elston2005zooming} \textcolor{black}{designed a semantic priming experiment in which bilingual German-English participants read English sentences that concluded with either an interlingual homograph or a neutral control word.} After each sentence, they completed a lexical decision task, where they were shown a target word and asked to judge whether it was a real word. Some target words were semantically linked to the homograph’s meaning in German (their L1), to test for native language influence. Additionally, half of the participants were made to watch a German-language film prior to the experiment to further activate their L1 semantic context. Reaction times and event-related potentials (ERPs) were measured for all participants to capture both behavioural and neural responses to the overall task. The results showed participants to respond much more slowly when the target words were related to the L1 meaning of the presented homograph. Their ERP data revealed that targets related to the L1 meaning of the homograph elicited delayed responses, indicating greater semantic conflict during lexical processing. These findings strongly support the view that interlingual homographs co-activate both L1 and L2 meanings, even when surrounded by an entirely L2 context. Moreover, recent activation of L1 (by watching the German film before the experiment) increased the interference from the L1 meaning, thereby indicating the influence of global language context on bilingual processing. 

\textcolor{black}{Recently, there has been a growing interest in applying psychological and linguistic tests to evaluate and understand the intricacies of generative AI systems.~\citet{Jin2024LanguageMA} used the Trolley problem~\cite{thomson1984trolley} to analyse the moral alignment of multilingual models, finding an absence of coherent moral reasoning across languages. Similarly,~\citet{arnett2025acquisition} adapted structural priming tests, which have been integrated into grammatical representation in the human mind, to uncover the multilingual LLM's ability to acquire a shared grammatical representation of language during studies. Human language grounding studies have also been utilised as tools in understanding LLMs~\cite{10.1162/coli_a_00531,Apidianaki2024LanguageLR}. In our analysis of the bilingual processing abilities of multilingual LLMs, we take inspiration from these studies and interpret our findings with the help of the BIA+ model. Apart from being a theoretically robust explanation of human bilingual processing, this model asserts that the mental lexicon of bilingual speakers is integrated. This means that at any given moment, a bilingual speaker has access to linguistic information from every language source that exists in their integrated lexicon. Similarly, the model posits that depending on the nature of a demand, bilingual speakers can sift through the integrated lexicon and retrieve linguistic information from a specific language source. We find this to be the most compelling theory to analyse and predict the performance of multilingual LLMs on bilingual processing tasks. Further, the BIA+ model is rooted in psycholinguistics and cognitive science. Thus, by using this model to analyse how multilingual LLMs process cognates, non-cognates, and interlingual homographs, we aim to make our findings grounded in theories of human cognition that can ultimately develop a robust explanation of bilingual processing as a computational process.}

\section{Materials and Methods}\label{sec:material}

\subsection{Background}

\textbf{Large Language Models} are transformer architecture-based~\cite{vaswani2017attention} auto-regressive models that are trained to predict the likelihood of a token sequence. Given a sequence of tokens $x=(x_1,x_2,...,x_t-1)$,  the LLM learns to predict the next token $x_t$ by learning the conditional entropy:
\begin{equation}
P(x_t \mid x_1, x_2, \dots, x_{t-1}; \theta)
\end{equation}
where $x_{<t}$ refers to the sequence $(x_1, \dots, x_{t-1})$, and $\theta$ are the model parameters. The LLM is trained using maximum likelihood estimation by minimising the negative log-likelihood~\cite{bishop2006pattern}) over a training corpus:
\begin{equation}
\mathcal{L}(\theta) = - \sum_{t=1}^{T} \log P(x_t \mid x_{<t}; \theta)
\end{equation}
After this training stage, the LLM is aligned with human preferences through \textit{ reinforcement learning with human feedback}~\cite{ouyang2022training}. In this stage, a reward model is trained using human comparisons to the model's outputs to prefer the outputs generated by humans. The LLM is then fine-tuned using Proximal Policy Optimisation (PPO) to maximise expected reward:
\begin{equation}
\mathcal{L}_{\text{RL}}(\theta) = \mathbb{E}_{x \sim \pi_\theta} \left[ r(x) \right]
\end{equation}
where $r(x)$ is the scalar reward assigned by the reward model. This results in the LLM generation being more aligned with human preferences.

During inference, given a context, the LLM generates a sequence by selecting one token at a time from the predicted distribution. We use \textit{greedy decoding} in our experiments, where at each time step $t$, the next token $\hat{x}_t$ is selected as the one with the highest probability:
\begin{equation}
\hat{x}_t = \arg\max_{x \in \mathcal{V}} P(x \mid \hat{x}_{<t})
\end{equation}
where $\mathcal{V}$ is the vocabulary of the model and $\hat{x}_{<t}$ denotes the previously generated tokens. \textcolor{black}{We use greedy decoding as it is deterministic and easily reproducible. At each step of generation, we record whether the correct token with respect to the task is present in the model’s prediction (marked as true) or not (marked as false).}

\subsection{Dataset}
\label{sec:materical_n_methods}

\paragraph{\bf Word lists} We created three novel datasets of cognates, non-cognates, and interlingual homographs between English-Spanish, English-French, and English-German. Each dataset consisted of $420$ word pairs; these pairs were equally split into noun pairs and adjective pairs.

Each experimental item, across all word types and languages, was curated and verified by linguists with the help of the Merriam-Webster Dictionary (English)~\footnote{Merriam-Webster.com Dictionary, accessed August 2024, https://www.merriam-webster.com.}, Real Academia (Spanish)~\footnote{Real Academia Española, \textit{Diccionario de la lengua española}, accessed August 2024, https://www.rae.es/.}, Reverso Dictionary (French)~\footnote{\textit{Reverso Dictionary}, accessed August 2024, https://dictionary.reverso.net/.}, and Digital Dictionary of the German Language (German)~\footnote{\textit{Digitales Wörterbuch der deutschen Sprache (DWDS)}, accessed August 2024, https://www.dwds.de/.}. Specifically, we manually extracted and curated English-Spanish, English-French, and English-German cognate pairs by referring to \emph{NTC's Dictionary of Spanish Cognates}~\cite{nash1993ntc}, \emph{French and English Cognates} listed by Rigdon~\cite{rigdon2017french}, and \emph{English / German Cognates: English / Deutsch 
% \todo{why all frist letters are caption?} Names of books so I emp them
Cognates}~\cite{rigdon2018english} in Spanish, French, and German, respectively. \textcolor{black}{While selecting cognate words, our primary objective was to focus on words that had a common etymological root, with similar orthography and meaning across two languages. However, during the extraction process, we also came across word pairs that share the same meaning in some, but not all, contexts. For example, the word \textit{"police"} in French means the same thing as in English. However, depending on the context, it can also be used to refer to a font type or a policy. Such pairs were classified as \textit{partial cognates} and were excluded from our dataset.}

Interlingual homographs across all languages were similarly manually extracted and curated from \emph{NTC's Dictionary of Spanish False Cognates (Spanish)}~\cite{ntc_spanish}, \emph{NTC's Dictionary of Faux Amis (French)}~\cite{ntc_french}, and \emph{NTC's Dictionary of German False Cognates (German)}~\cite{ntc_german}. \textcolor{black}{Our primary objective in the selection process was to focus on homograph pairs that were orthographically as similar as possible, while having drastically different meanings across two languages.} Their non-cognate translations were derived from Real Academia (Spanish), Reverso Dictionary (French), and Digital Dictionary of the German Language (German). These translations constituted our set of non-cognate words.  \textcolor{black}{Once the three datasets (i.e., English-Spanish, English-French, and English-German) were finalised, each of them was split up to $20\%$ to form the training set that is used by us in formulating the examples for in-context learning}.

\begin{table*}[!ht]
\centering
\resizebox{0.8\textwidth}{!}{ 
\begin{tabular}{lcccccc}
\hline
\backslashbox{Type}{Model} & BLOOMZ& LLaMA2 & LLaMA3 & LLaMA3.1 & MISTRAL & AVG.\\ \hline
\multicolumn{7}{c}{\cellcolor[HTML]{EEEEEE}English-German} \\ \hline

\hline
Cognates& 0.318 & 0.310 & 0.305 & 0.305 & 0.280 & 0.304 \\
Homographs& 0.329 & 0.318 & 0.319 & 0.319 & 0.302 & 0.317 \\
Non-cognates & 0.013 & 0.015 & 0.019 & 0.019 & 0.009 & 0.015 \\
\hline
\multicolumn{7}{c}{\cellcolor[HTML]{EEEEEE}English-Spanish} \\ \hline

\hline
Cognates&0.238 & 0.217 & 0.229 & 0.229 & 0.190 & 0.221 \\
Homographs& 0.410 & 0.392 & 0.407 & 0.407 & 0.380 & 0.399 \\
Non-cognates & 0.014 & 0.005 & 0.013 & 0.013 & 0.008 & 0.011 \\
\hline
\multicolumn{7}{c}{\cellcolor[HTML]{EEEEEE}English-French} \\ \hline

\hline
Cognates& 0.293 & 0.258 & 0.272 & 0.272 & 0.228 & 0.265 \\
Homographs&0.377 & 0.365 & 0.386 & 0.386 & 0.353 & 0.373 \\
Non-cognates & 0.025 & 0.027 & 0.027 & 0.027 & 0.024 & 0.026 \\
\hline
\end{tabular}
}
\caption{Token overlap between cognates, homographs and non-cognates for the five models under our consideration.}
\vspace{-5mm}
\label{tab:overlap}
\end{table*}

\paragraph{\bf Semantic constraint sentences} For our final experiment, we were required to generate low- and high-semantic constraint sentences for homograph pairs in the English-Spanish, English-French, and English-German datasets. \textcolor{black}{As explained by \citet{dijkstra2015sentence},  sentences with low semantic constraint are those where the meaning of the initial words of the sentence does not strongly predict the remaining words of the sentence (e.g., \textit{"My sister was hungry and took the last ....}", where there are multiple possibilities about what the missing word could be, such as \textit{"apple"}, \textit{"banana"}, \textit{"slice"}, etc.). In contrast, sentences with high semantic constraint are those where the meaning of the initial words of the sentence definitively predicts the remaining words (e.g., \textit{"The honeybee drinks nectar from the ...."}, where there is a strong probability of the missing word being \textit{"flower"})}. Crucially, we required such sentences in non-English languages as well, to test the influence of language context on bilingual processing in LLMs.

We resorted to the automatic generation of these sentences with the help of GPT-4~\cite{achiam2023gpt}. This was achieved by first prompting the LLM to create sentences with low- and high-semantic constraint in English for each homograph in the three datasets. The generated sentences were then curated and verified by linguists before being finalised for further experimentation. Similarly, the LLM was prompted to generate low- and high-semantic constraint sentences in English, but with the non-English meaning of each homograph as the target word. After undergoing the same curation process, these sentences were translated into Spanish, French, or German, depending on the homograph pair. The Google Translate software assisted us in the translation procedure. Once all English and non-English sentences had been finalised, the target homograph in each sentence was replaced with a $\langle MASK \rangle$ token, allowing it to be substituted with its corresponding pair in another language. This was in accordance with the design of our third experimental task. This creates our corpus of $1680$ sentences. 
\textcolor{black}{Examples \hyperref[example:task1]{A}, \hyperref[example:task2]{B}, and \hyperref[example:task3]{C} show the prompts used in our experiment. We discuss these prompts later in detail.}

\subsection{Token Overlap Analysis Across Models}
\textcolor{black}{To quantify how different LLMs represent words at the sub-word level, we perform a token overlap analysis on word pairs in our dataset, focusing on \textbf{cognates}, \textbf{non-cognates}, and \textbf{interlingual homographs} for each of the models used in our study.}

\textcolor{black}{For a word pair $(w_1, w_2)$, let $\text{tokens}_1$ and $\text{tokens}_2$ denote the sets of tokens obtained. The \textbf{token overlap} was defined as: $\text{overlap}(w_1, w_2) = \frac{|\text{tokens}_1 \cap \text{tokens}_2|}{|\text{tokens}_1 \cup \text{tokens}_2|}$}

\textcolor{black}{This yields a value between 0 and 1, where 1 indicates complete overlap and 0 indicates no shared tokens. Table~\ref {tab:overlap} shows the average overlap score for the three word groups using the five models used in our experiment. We note that cognates and homographs have 18 times higher overlap than non-cognates.}

\subsection{Experimental Setup}

We experiment with \textcolor{black}{five publicly available multilingual LLMs: BLOOMZ~\cite{workshop2023bloom176bparameteropenaccessmultilingual}\footnote{Hugging Face identifiers: \texttt{bigscience/bloomz-7b1}}, Mistral~\cite{barker2023travelingwavesreactiondiffusionequations}\footnote{Hugging Face identifiers: \texttt{mistralai/Mistral-7B-Instruct-v0.2}}, LLaMA-2~\cite{touvron2023LLaMA2openfoundation}\footnote{Hugging Face identifiers: \texttt{meta-LLaMA/LLaMA-2-7b-chat-hf}}, LLaMA-3~\cite{LLaMA3}\footnote{Hugging Face identifiers: \texttt{meta-LLaMA/Meta-LLaMA-3-8B-Instruct}}, and LLaMA-3.1~\cite{LLaMA3}\footnote{Hugging Face identifiers: \texttt{meta-LLaMA/Meta-LLaMA-3.1-8B-Instruct}}}. We investigate their ability to process and interpret three types of words -- cognates, non-cognates, and interlingual homographs-- across three language pairs -- English-Spanish, English-German, and English-French. 
%To create our dataset, we utilise linguistic word lists corresponding to these three-word types and extracted nouns and adjectives in the aforementioned language pairs. Our dataset consists of 140 word pairs for each category in each language pair, totalling 1,260-word pairs. Additionally, we utilise language-specific dictionaries to extract and verify the (dis)similarity in meaning of the word pairs. 

Recent studies~\cite{wendler-etal-2024-LLaMAs,zhao2024large,tang-etal-2024-language} suggest that multilingual models predominantly rely on English while performing reasoning tasks across languages. They tend to translate non-English inputs into English in the initial layers, perform reasoning in English in the intermediate layers, and finally translate the output back to the target language in the later layers. This internal pivot to English is attributed to the predominance of English in the training data of many models. This is similar to the concept of first language (L1) proficiency in dealing with human cognition. Thus, in our experimental tasks, we utilise English as the dominant language for these models. Significance testing of our results can be found in \hyperref[SI:sub_section1]{Appendix A}.

\section{Experimental Results}

\begin{figure*}[t!]
\centering
\includegraphics[width=\linewidth]{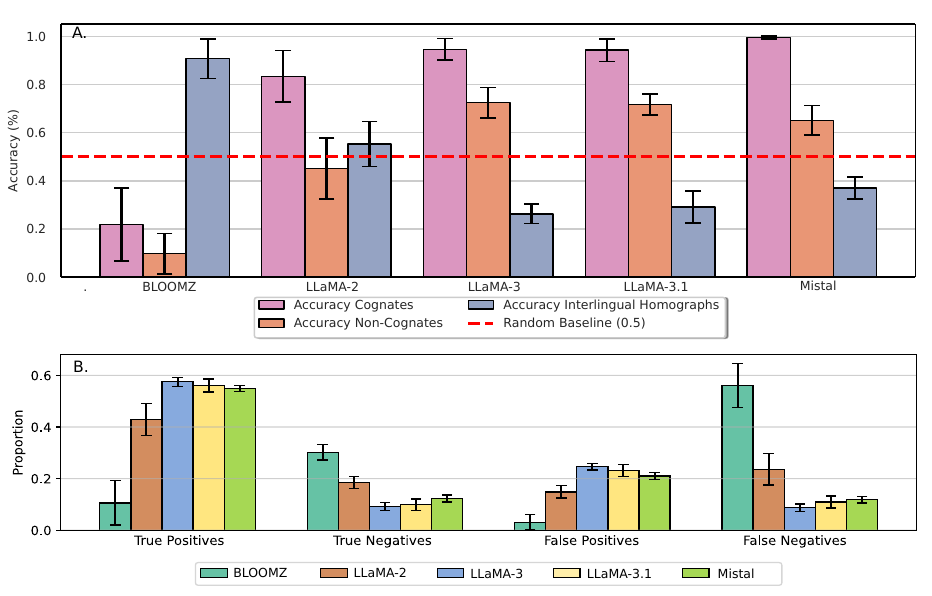}
\caption{\textbf{(A) Word pair disambiguation accuracy.} The figure shows the average performance of five multilingual LLMs \textcolor{black}{(the error bar notes the standard deviation over five runs.)} in identifying word pairs to have the same meaning or not. The experiments are conducted on five random seed values, and the variations in performance on these runs are represented by error bars. All the models seem to perform better on cognates than non-cognate pairs, highlighting the role of cognate facilitation in LLMs. All models except BLOOMZ seem to perform poorly in disambiguating phonograph pairs compared to cognates, further showing their utilisation of orthographical signals in the disambiguation task. \textbf{(B) Class distribution of prediction for word pair disambiguation task.} For each of the five models, we show the proportion of true positives, true negatives, false positives, and false negatives, averaged across all seeds (the error bar notes the standard deviation). BLOOMZ shows a higher proportion of false negatives compared to other models. We attribute this to the anomalous performance of BLOOMZ.}
\label{fig:main_t1}
\vspace{-5mm}
\end{figure*}

\subsection{Cognate, Non-cognate and Homograph Disambiguation} Studies on bilingualism show that word retrieval improves when both languages share an integrated lexicon and representation space~\cite{dijkstra2005bilingual}. With respect to LLMs, multilingual representation is aligned by leveraging cross-lingual anchors. This creates shared word representation spaces for common or similar words across multiple languages~\cite{hua-etal-2024-mothello,Deshpande2021WhenIB} and enables cross-lingual processing and performance in LLMs during the early stages of their training~\cite{wang-etal-2024-probing-emergence}. 
% However, these shared sub-spaces can be bifurcated into language-specific regions~\citet{Chang2022TheGO}, indicating that certain axes of the hidden states of the model encode language-sensitive information of a sentence. This axis can be used by the model to identify the language of the words, and may implicitly encode further information about the language. 
Hence, our first experimental task aims to analyse whether multilingual LLMs can implicitly separate cognates, non-cognates, and interlingual homographs. The primary objective behind our task design is to understand how accurately semantic and language information encoded in the representation spaces helps in the disambiguation of different types of words.

\paragraph{\bf Task description} Our prompt consists of a task instruction designed to guide the models in performing the word pair disambiguation. A demonstration of labelled examples follows the instruction, and finally, the word pair we want the model to evaluate, as shown in Example A. The model is instructed to label the pair as \emph{True} if the words have the same meaning, or \emph{False} if they have different meanings. To create the demonstration, we sample four examples of word pairs with the same meaning and four examples of word pairs with different meanings. This is done to reduce any chance of bias being induced in the model by the demonstrations (refer to \hyperref[SI:sub_section2]{Appendix B} for analysis on varying numbers of examples). Furthermore, we run our experiments on five different samples of demonstrations to further get a generalised view.

\begin{tcolorbox}[
colframe=black!50!black, 
colback=black!10, 
title=Example A: Disambiguation task example,
label=example:task1]
Given an English-French word pair
label them \emph{True} if they have the
same meaning, else label them as
\emph{False}.

Following are a few labelled examples
to assist you with this task.

Land and Lande $\rightarrow$ \textit{False}

Real and Réel $\rightarrow$ \textit{True}

$\cdots$

Sensitive and Sensitive $\rightarrow$ \textit{False}

Medicine and Médecin $\rightarrow$ \textit{False}

Now label the following.

Park and Parc $\rightarrow$
\end{tcolorbox}

 \noindent \textbf{Can multilingual models disambiguate cognates and homographs?} Figure~\hyperref[fig:main_t1]{2(A)} shows the class-wise accuracy of the multilingual models in disambiguating between cognates, non-cognates, and interlingual homographs word pairs. Mistral, LLaMA-3 and LLaMA-3.1 have an impressive ability to disambiguate cognates, with accuracies of 99.44\%, 94.58\%, and 94.23\%, respectively, and for non-cognates, with accuracies of 65.11\%, 72.48\%, and 71.62\%. However, they struggle to disambiguate interlingual homograph pairs, showing below-random performance with accuracy of 37.04\%, 24.86\%, and 29.12\%, respectively. On the other hand, BLOOMZ can classify interlingual homographs, with an accuracy of 90\%, much better than the aforementioned models, but displays dismal performance for cognates and non-cognate word pairs, 21.90\% and 9.82\% respectively. Notably, LLaMA-2 shows above-random performance for interlingual homographs, with an accuracy of 55.13\% while also performing well for cognates with an accuracy of 83\%, indicating a slight improvement in handling this challenging class (refer to the \hyperref[SI:sub_section3]{Appendix C}, for an analysis of each language pair).

To further analyse the unique ability of BLOOMZ \textcolor{black}{to classify interlingual homographs, we investigate the prediction of the models with ground truth. We note that BLOOMZ disproportionally labels word pairs as semantically dissimilar compared to any other model. This is evident from the class distribution plots, as seen in Figure~\hyperref[fig:main_t1]{2(B)}}. This indicates that BLOOMZ is highly biased in favour of predicting \emph{False}, thus leading to its unique results.

\begin{table*}[!ht]
\centering
\resizebox{\textwidth}{!}{ 
\begin{tabular}{l|ccccc|ccccc}
\hline
\backslashbox{Type}{Model} & BLOOMZ& LLaMA2 & LLaMA3 & LLaMA3.1 & MISTRAL & BLOOMZ& LLaMA2 & LLaMA3 & LLaMA3.1 & MISTRAL \\ \hline
\multicolumn{1}{c}{\cellcolor[HTML]{EEEEEE}} &
\multicolumn{10}{c}{\cellcolor[HTML]{EEEEEE}English-French} \\ \hline
\multicolumn{1}{c}{\cellcolor[HTML]{EEEEEE}} &
\multicolumn{5}{c}{\cellcolor[HTML]{EEEEEE}Identical} & \multicolumn{5}{c}{\cellcolor[HTML]{EEEEEE}partially identical} \\
\hline
Cognates&4.30&100.00&100.00&100.00&100.00&12.70&86.30&99.09&97.27&100.00\\
Homographs&100.00&0.00&0.00&6.00&8.80&100.00&65.00&27.00&38.04&43.00\\ \hline
\multicolumn{1}{c}{\cellcolor[HTML]{EEEEEE}} &
\multicolumn{10}{c}{\cellcolor[HTML]{EEEEEE}English-German} \\ \hline
\multicolumn{1}{c}{\cellcolor[HTML]{EEEEEE}} &
\multicolumn{5}{c}{\cellcolor[HTML]{EEEEEE}Identical} & \multicolumn{5}{c}{\cellcolor[HTML]{EEEEEE}partially identical} \\
\hline
Cognates&0.70&100.00&100.00&100.00&100.00&8.40&89.90&97.47&97.47&100.00\\
Homographs&0.97&2.90&2.90&2.90&8.80&98.90&71.71&32.30&24.24&33.33\\ \hline

\multicolumn{1}{c}{\cellcolor[HTML]{EEEEEE}} &
\multicolumn{10}{c}{\cellcolor[HTML]{EEEEEE}English-Spanish} \\ \hline
\multicolumn{1}{c}{\cellcolor[HTML]{EEEEEE}} &
\multicolumn{5}{c}{\cellcolor[HTML]{EEEEEE}Identical} & \multicolumn{5}{c}{\cellcolor[HTML]{EEEEEE}partially identical} \\
\hline
Cognates&0.00&100.00&100.00&100.00&100.00&42.42&85.85&98.98&97.97&97.97\\
Homographs&0.96&0.00&0.00&0.00&3.80&81.30&85.04&40.18&38.31&57.94\\ \hline
\end{tabular}
}
\caption{Accuracy for word disambiguation task to compare the difference in identical and partially identical words.}
\vspace{-5mm}
\label{tab:complete-partial}
\end{table*}

\textcolor{black}{\paragraph{ \bf Completely and Partially Identical Cognates:} Our results also point to the role of cognate facilitation in LLMs' bilingual word processing ability (as evident in Figure~\hyperref[fig:main_t1]{2(A)} and \hyperref[SI:sub_section3]{Appendix C}). Orthographic similarities make it much easier for all models to recognise and classify cognate words, as compared to non-cognates. Further, the orthographic similarity of interlingual homographs makes it difficult for the model to recognise and access their distinct meanings. To further analyse this, we divide our dataset of interlingual homographs and cognates into orthographically completely identical word pairs (e.g., the English-German homograph pair \emph{`gift'}, which refers to \emph{a present} or \emph{an offering} in English, while \emph{poison} in German), and orthographic partially identical pairs (e.g., the English-Spanish homograph pair \emph{`blank-blanco'}, where \emph{`blank'} refers to \emph{a space/gap} while \emph{`blanco'} refers to \emph{a goal, target, or aim}). With this manipulation, we could factor in the influence of orthographic similarity on bilingual word recognition in these models. From a linguistic perspective, a word with many other similar words within the same language or even cross-linguistically is said to have a dense neighbourhood. The similarity between these words could be orthographic, phonological, or a combination of both of these elements. This neighbourhood density plays a crucial role in human bilingual processing by influencing the level of activation that the target word receives to be selected and identified~\cite{dijkstra2002architecture}. Typically, words with high neighbourhood density are more difficult to recognise, as compared to words with a sparse neighbourhood~\cite{van1998orthographic}. In a similar vein, our findings note that LLaMA-2, LLaMA-3, LLaMA-3.1, and Mistral perform 5\% better while disambiguating identical cognates, as compared to their not completely identical counterparts (see Table~\ref{tab:complete-partial}). Further, the performance of these models while disambiguating partially identical homographs is $15\times$ better than when they process identical interlingual homographs (see Table~\ref{tab:complete-partial}). }

% \begin{figure*}[t!]
% \centering
% \includegraphics[width=\linewidth]{fig/main2.pdf}
% \caption{Placeholder image of a frog with a long example legend to show justification setting.}
% \label{fig:main_t2}
% \end{figure*}

\subsection{Semantic Judgment} Previous experimental results showcase multilingual LLMs' inability to disambiguate interlingual homographs and further show the utility of orthographical signals in word pair disambiguation. However, these findings offer little insight into the models' ability to store, retrieve and understand the distinct semantic links associated with the word pairs. LLMs' training objective doesn't explicitly prod the models to learn linguistic and semantic features associated with a word. LLMs are trained to capture statistical distributions of the next token prediction task. Hence, understanding the meaning of the words is not an explicit objective of their training. This has led to some studies calling them "stochastic parrots''~\cite{10.1145/3442188.3445922,10.1162/coli_a_00522}. Studies have shown their lack of comprehension in basic tasks associated with understanding a word like \textcolor{black}{counting the number of occurrences of a letter in a word~\cite{shin2024large,doi:10.1073/pnas.2215907120}. This phenomenon can be attributed to the process of tokenisation, where words are divided into sub-words rather than individual letters. These sub-words contain the orthographical cues utilised by an LLM in word processing. Studies on prompt probing have shown their ability to encode linguistic features like part-of-speech tagging, named-entity recognition, sentiment analysis, etc., internally in their hidden states~\cite{gurnee2023finding,Galke2024} related to the words and sub-words they process.} Based on these studies and to address the gap in the first disambiguation task, we find it fitting to analyse these models' semantic storage and retrieval ability. Specifically, we study their ability to retrieve semantic information about cognates, non-cognates and interlingual homographs as they pose a special challenge, and further analyse any dependency between their semantic judgment and performance in the disambiguation task.

\noindent \textbf{Task description.} To analyse the semantic retrieval capabilities of multilingual LLMs, we evaluate their ability to select the correct meaning of a word. The model is prompted with two possible meanings for a given word: one correct and one incorrect. We then assess its ability to accurately identify the correct meaning.

For each word in our dataset, we construct a correct meaning corpus by manually extracting meaning from language-specific dictionaries (see Section \ref{sec:material} for details). To sample the incorrect meaning:
\begin{itemize}
    \item For \emph{homographs}, the meaning of the corresponding pair is used as the incorrect sample.
    \item For \emph{cognates} and \emph{non-cognates}, a semantically dissimilar meaning is sampled from our corpus using the sentence transformer~\cite{reimers-2019-sentence-bert}. Specifically, we select a word meaning with a cosine similarity below $0.6$ as the incorrect sample.
\end{itemize}
\textcolor{black}{The model is prompted with a task instruction followed by four labelled examples to perform the semantic judgment, as demonstrated in Example B.}\\

\begin{tcolorbox}[
colframe=black!50!black, 
colback=black!10,
title=Example B: Semantic judgment task example,
label=example:task2
% breakable,
]
Given a word and two options, select the correct meaning of the word from the two provided options.

Following are a few labelled examples to assist you with this task.

What is the meaning of ``correcto'' in Spanish?

\begin{enumerate}
    \item Free of errors or defects, $\ldots$
\end{enumerate}

Now label the following

What is the meaning of ``embarazada'' in Spanish?

\begin{enumerate}
    \item Said of a woman who has conceived and has the fetus or child in her womb.
    \item feeling or showing a state of self-conscious confusion and distress.
\end{enumerate}
\end{tcolorbox}
\begin{table*}[!h]
\centering
\resizebox{\textwidth}{!}{ 
\begin{tabular}{l|ccccc|ccccc}
\hline
\backslashbox{Type}{Model} & BLOOMZ & LLaMA-2 & LLaMA-3 & LLaMA-3.1 & MISTRAL & BLOOM & LLaMA-2 & LLaMA-3 & LLaMA-3.1 & MISTRAL \\ \hline
\multicolumn{1}{c}{\cellcolor[HTML]{EEEEEE}} &
\multicolumn{5}{c}{\cellcolor[HTML]{EEEEEE}English} & \multicolumn{5}{c}{\cellcolor[HTML]{EEEEEE}French} \\ \hline
Cognates      & \cellcolor{yellow!60}$50.73_{0.01}$ & \cellcolor{red!40}$18.20_{0.10}$ & \cellcolor{green!40}$57.19_{0.16}$ & \cellcolor{green!40}$57.39_{0.17}$ & \cellcolor{green!40}$57.19_{0.17}$ & \cellcolor{yellow!50}$45.11_{0.02}$ & \cellcolor{green!50}$69.92_{0.18}$ & \cellcolor{green!40}$57.44_{0.17}$ & \cellcolor{green!60}$76.99_{0.10}$ & \cellcolor{green!50}$65.11_{0.17}$ \\
Non-cognates  & \cellcolor{yellow!50}$47.27_{0.01}$ & \cellcolor{red!40}$20.10_{0.13}$ & \cellcolor{green!40}$56.69_{0.09}$ & \cellcolor{green!40}$58.15_{0.14}$ & \cellcolor{green!40}$57.79_{0.01}$ & \cellcolor{green!40}$59.40_{0.02}$ & \cellcolor{green!50}$69.47_{0.03}$ & \cellcolor{green!40}$58.65_{0.21}$ & \cellcolor{green!60}$76.09_{0.09}$ & \cellcolor{green!40}$63.76_{0.13}$ \\
Homographs    & \cellcolor{yellow!50}$45.86_{0.00}$ & \cellcolor{red!40}$21.55_{0.11}$ & \cellcolor{green!30}$55.14_{0.16}$ & \cellcolor{yellow!50}$53.83_{0.13}$ & \cellcolor{green!40}$58.05_{0.13}$ & \cellcolor{green!30}$56.69_{0.13}$ & \cellcolor{green!50}$65.41_{0.14}$ & \cellcolor{yellow!50}$53.23_{0.13}$ & \cellcolor{green!50}$69.62_{0.12}$ & \cellcolor{yellow!50}$54.29_{0.13}$ \\ \hline \hline
Average & \cellcolor{yellow!50}$47.95$& \cellcolor{red!40}$19.95$&\cellcolor{green!30}$56.34$&\cellcolor{green!30}$56.45$&\cellcolor{green!30}$57.67$&\cellcolor{yellow!50}$53.77$&\cellcolor{green!50}$68.26$&\cellcolor{green!30}$56.44$&\cellcolor{green!60}$74.2$&\cellcolor{green!40}$61.05$\\ \hline \hline
\multicolumn{1}{c}{\cellcolor[HTML]{EEEEEE}} &
\multicolumn{5}{c}{\cellcolor[HTML]{EEEEEE}German} & \multicolumn{5}{c}{\cellcolor[HTML]{EEEEEE}Spanish} \\ \hline 
Cognates      & \cellcolor{yellow!50}$44.21_{0.05}$ & \cellcolor{yellow!50}$41.80_{0.12}$ & \cellcolor{green!40}$56.39_{0.15}$ & \cellcolor{green!40}$57.64_{0.10}$ & \cellcolor{green!40}$59.02_{0.15}$ & \cellcolor{yellow!50}$43.16_{0.02}$ & \cellcolor{green!40}$61.20_{0.09}$ & \cellcolor{green!40}$58.20_{0.16}$ & \cellcolor{green!50}$73.53_{0.12}$ & \cellcolor{green!40}$65.26_{0.17}$ \\
Non-cognates  & \cellcolor{green!40}$58.65_{0.02}$ & \cellcolor{yellow!50}$42.26_{0.11}$ & \cellcolor{green!40}$57.44_{0.10}$ & \cellcolor{green!40}$57.94_{0.14}$ & \cellcolor{green!40}$56.92_{0.02}$ & \cellcolor{green!40}$60.15_{0.01}$ & \cellcolor{green!40}$62.26_{0.16}$ & \cellcolor{green!40}$62.26_{0.13}$ & \cellcolor{green!50}$74.29_{0.12}$ & \cellcolor{green!40}$65.56_{0.13}$ \\
Homographs    & \cellcolor{green!40}$60.00_{0.03}$ & \cellcolor{yellow!50}$44.81_{0.16}$ & \cellcolor{yellow!50}$54.88_{0.01}$ & \cellcolor{green!40}$56.32_{0.02}$ & \cellcolor{yellow!50}$54.89_{0.02}$ & \cellcolor{green!40}$58.05_{0.02}$ & \cellcolor{green!40}$63.61_{0.09}$ & \cellcolor{yellow!50}$51.13_{0.03}$ & \cellcolor{green!50}$72.33_{0.06}$ & \cellcolor{green!40}$63.31_{0.06}$ \\
\hline \hline
Average & \cellcolor{green!30}$54.28$& \cellcolor{yellow!50}$42.95$&\cellcolor{green!30}$56.23$&\cellcolor{green!30}$57.3$&\cellcolor{green!30}$56.94$&\cellcolor{yellow!50}$53.78$&\cellcolor{green!40}$62.35$&\cellcolor{green!30}$57.19$&\cellcolor{green!60}$73.38$&\cellcolor{green!40}$64.71$\\ \hline \hline
\end{tabular}
}
\caption{Performance of different language models (BLOOM, LLaMA-2, LLaMA-3, LLaMA-3.1, and MISTRAL), \textcolor{black}{along with their standard deviation}, across linguistic types (cognates, non-cognates, homographs) on the semantic judgment task in four languages (English, French, German, and Spanish). Scores are accuracy-based and colour-coded to indicate performance -- darker green represents higher scores, yellow represents mid-range scores, and red represents lower scores. An average row for each language block highlights the overall performance trend per model for the language.}
\label{tab:semantic_judgement}
\vspace{-5mm}
\end{table*}

\noindent \textbf{Can multilingual models understand a word?} \textcolor{black}{Table~\ref{tab:semantic_judgement} shows the performance of five models in our semantic judgment task across four languages.} The overall performance of LLaMA-3.1, Mistral, LLaMA-3, BLOOMZ, and LLaMA-2 is 65.33\%, 60.09\%, 56.55\%, 52.44\% and 48.37\%, respectively. Among these, Mistal and LLaMA-3.1 perform considerably better than the other models, showcasing their superior ability to handle semantic judgment tasks. However, on further inspection, we observe that for most cases, all models struggle to achieve a performance above 65\%, highlighting the inherent difficulty in retrieving the semantic meaning of words.

Interestingly, if we analyse the models' performance on word type cognates, non-cognates and interlingual homographs, we do not find a large variation in performance. The overall performance of the models on cognates, non-cognates and homographs is 69.73\%, 72.75\% and 69.56\%, respectively. This suggests that no single category is easier for the models to retrieve the meaning of, which is contrary to the expectation of cognate facilitation. This also contradicts our findings from the word-pair disambiguation task, where cognate facilitation was evident, suggesting that while models utilise orthographic signals in the disambiguation task, this facilitation does not help them in retrieving word meanings.

To further analyse if there is any correlation between the semantic judgment capacity of LLMs and their word pair disambiguation ability, we analyse the conditional entropy between the two tasks. Conditional entropy, $H(Y|X)$, measures the uncertainty in a random variable $Y$ given that the value of the random variable $X$ is known. If $H(Y|X)$ is similar to $H(Y)$, the entropy of $Y$, two random variables are considered to be independent (see \hyperref[SI:sub_section4]{Appendix D} for details).

We use the LLMs' output for the semantic judgment task of both words in a pair to determine whether the model understands the meaning of both words, only one, or neither. These labels, along with the model's prediction for the word pair disambiguation task, were used to determine correlation. We note $H(\text{Semantic Judgment}\mid\text{Word Pair Disambiguation})\sim H(\text{Semantic Judgment})$ and $H(\text{Word Pair Disambiguation}\mid\text{Semantic Judgment}) \sim H(\text{Word Pair Disambiguation})$, indicating the independence of the two variables (see~\hyperref[SI:sub_section4]{Appendix D}, Table~\ref{tab:entropy}). These results suggest that the model relies primarily on orthographic cues when processing word pairs in isolation, rather than leveraging semantic understanding.

 Given the training dynamic of LLMs, recent studies~\cite{10.1162/coli_a_00522,ref1,pavlick2023symbols,mollo2023vector} argued that LLMs can't establish the \textit{word-to-world} connection needed for meaningful human language understanding as they are trained to imitate patterns without any explicit grounding in reference. Our results support this argument. The lack of utility of semantic cues in the disambiguation task and the lack of utility of orthographic signals in semantic judgment may be caused by the lack of grounding in the LLM training regime. This inherently can cause difficulty for LLMs to process words in isolation, as reflected in our experimental tasks thus far. 

 \begin{table*}[!ht]
\centering
\resizebox{\textwidth}{!}{ 
\begin{tabular}{l|ccccc|ccccc}
\hline
\backslashbox{Type}{Model} & BLOOMZ & LLaMA2 & LLaMA3 & LLaMA3.1 & MISTRAL & BLOOMZ & LLaMA2 & LLaMA3 & LLaMA3.1 & MISTRAL \\ \hline

\multicolumn{1}{c}{\cellcolor[HTML]{EEEEEE}} &
\multicolumn{10}{c}{\cellcolor[HTML]{EEEEEE}English} \\ \hline
\multicolumn{1}{c}{\cellcolor[HTML]{EEEEEE}} &
\multicolumn{5}{c}{\cellcolor[HTML]{EEEEEE}Identical} &
\multicolumn{5}{c}{\cellcolor[HTML]{EEEEEE}partially identical} \\ \hline
Cognates
&42.52&18.51&48.81&52.25&51.48
&51.75&19.61&58.01&57.56&57.87 \\
Homographs
&50.70&24.38&54.86&52.38&56.84
&44.23&22.36&55.20&54.57&58.51 \\ \hline

\multicolumn{1}{c}{\cellcolor[HTML]{EEEEEE}} &
\multicolumn{10}{c}{\cellcolor[HTML]{EEEEEE}French} \\ \hline
\multicolumn{1}{c}{\cellcolor[HTML]{EEEEEE}} &
\multicolumn{5}{c}{\cellcolor[HTML]{EEEEEE}Identical} &
\multicolumn{5}{c}{\cellcolor[HTML]{EEEEEE}partially identical} \\ \hline
Cognates
&50.43&65.22&57.39&77.39&64.35
&44.00&70.91&57.45&76.91&65.27 \\
Homographs
&60.61&64.85&48.48&67.27&44.85
&55.40&65.60&54.80&70.40&57.40 \\ \hline

\multicolumn{1}{c}{\cellcolor[HTML]{EEEEEE}} &
\multicolumn{10}{c}{\cellcolor[HTML]{EEEEEE}German} \\ \hline
\multicolumn{1}{c}{\cellcolor[HTML]{EEEEEE}} &
\multicolumn{5}{c}{\cellcolor[HTML]{EEEEEE}Identical} &
\multicolumn{5}{c}{\cellcolor[HTML]{EEEEEE}partially identical} \\ \hline
Cognates
&61.76&30.59&61.76&78.82&62.35
&38.18&45.66&60.40&71.52&66.46 \\
Homographs
&63.08&40.77&50.00&66.15&49.23
&59.25&45.79&54.58&69.35&60.00 \\ \hline

\multicolumn{1}{c}{\cellcolor[HTML]{EEEEEE}} &
\multicolumn{10}{c}{\cellcolor[HTML]{EEEEEE}Spanish} \\ \hline
\multicolumn{1}{c}{\cellcolor[HTML]{EEEEEE}} &
\multicolumn{5}{c}{\cellcolor[HTML]{EEEEEE}Identical} &
\multicolumn{5}{c}{\cellcolor[HTML]{EEEEEE}partially identical} \\ \hline
Cognates
&48.57&70.00&57.14&80.00&68.57
&42.35&61.68&57.14&72.77&64.87 \\
Homographs
&70.59&65.29&50.00&71.76&56.47
&53.54&64.24&54.34&70.30&65.66 \\ \hline

\end{tabular}
}
\caption{Accuracy for the semantic judgment task to compare the difference in identical and partially identical words.}
\vspace{-5mm}
\label{tab:complete-partial2}
\end{table*}

\textcolor{black}{\noindent\textbf{Completely and Partially Identical Cognates:}
Table~\ref{tab:complete-partial2} reports the accuracy of multilingual LLMs on the semantic judgment task, comparing performance on identical and partially identical cognates and homographs across four languages. In contrast to Table~\ref{tab:complete-partial}, where orthographic similarity played a key role in the word disambiguation task, we find no evidence that such cues facilitate semantic retrieval. Performance does not exhibit a consistent or coherent pattern across languages or model types for either fully identical or partially identical words. This absence of systematic effects further reinforces our finding that LLMs struggle to reliably link orthographic form with semantic representations.}

% \begin{figure*}[t!] % Use figure* to span both columns
%     \centering
%     \begin{minipage}{0.7\textwidth} % Width of the figure
%         \includegraphics[width=\linewidth]{fig/main1.pdf} % Include your figure
%     \end{minipage}%
%     \begin{minipage}{0.25\textwidth} % Width of the caption box
%         \captionof{figure}{Placeholder image of a frog with a long example legend to show justification setting. This caption can extend to multiple lines and will be properly justified.}
%         \label{fig:main_t1}
%     \end{minipage}
% \end{figure*}

% \begin{figure}%[tbhp]
% \centering
% \includegraphics[width=\linewidth]{fig/imbalance.pdf}
% \caption{Placeholder image of a frog with a long example legend to show justification setting.}
% \label{fig:imbalance_t1}
% \end{figure}
% \noindent 

\begin{figure*}[t!]
\centering
\includegraphics[width=\linewidth]{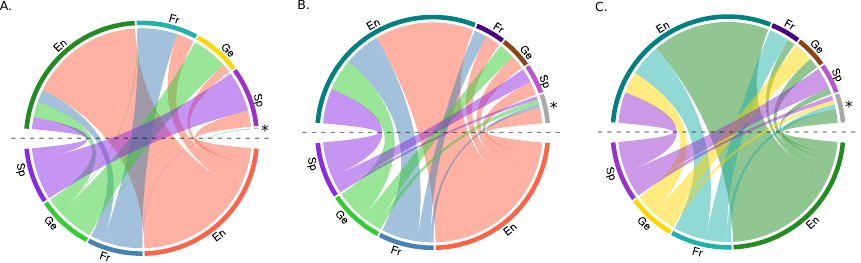}
\caption{\textbf{(A) Word language recognition.}  The top half shows the language predicted by the LLM; the bottom half shows the actual language. Chords indicate matches and mismatches. The model is correct 73\% of the time. In errors, it often defaults to the context language. Notably, some German words are misclassified as Dutch or French, despite the English context. \textbf{(B) Word meaning recognition.} The top half shows the language of the meaning used by the LLM; the bottom shows the word’s actual language. For many Spanish, German, and French words, the model uses English meanings, highlighting its reliance on context. This effect is weaker for English words, which are typically assigned English meanings regardless of context. \textbf{(C) Interplay between language and meaning.} The top half shows meaningful language; the bottom shows recognised word language. For non-English words, English meanings are often chosen, showing context sensitivity. For English words, this sensitivity is largely absent (En: English, Fr: French, Ge: German, Sp: Spanish, and *: Neither).}
\label{fig:main_t3_1}
\end{figure*}

\subsection{Semantic Constraint and Bilingual Word Processing in LLMs} 

So far, we have dealt with LLMs' word-processing ability in isolation. However, in natural conversation, words are always processed in the context of utterances~\cite{chomsky1953systems}. Previous studies by \citet{dijkstra2015sentence} showed that sentence characteristics such as the degree of semantic constraint and language of the sentence affect the processing of cognates by bilingual speakers. The semantic constraint, low or high, defines how strongly the meaning of the preceding sentence can determine the meaning of a target word. Identifying the language of the sentence provides an additional context that facilitates the prediction of the meaning of the target word, especially when it has a cognate status. 

% To explore this relation between word meaning and its context, we designed a task focused on interlingual homographs, specifically aiming to evaluate LLMs' disambiguation capabilities through semantic constraints. We created high- and low-semantic-constraint sentences for the interlingual homographs in our dataset. Further, in each sentence, we systematically replaced the interlingual homograph with its corresponding pair from the different languages, thus introducing cross-lingual ambiguity, leading to a dataset of 1680 sentences. We want to analyse the model's ability to deal with this ambiguity, by leveraging the semantic cues present in the sentence and the orthographical similarities that homograph pairs have.  

\textcolor{black}{We take inspiration from \citet{dijkstra2015sentence} to develop a novel third task that explores the LLMs' understanding of the relation between word meaning and its context.} We design a task focused on partly identical interlingual homographs, specifically aiming to evaluate LLMs' disambiguation capabilities through semantic constraints. 

We generate low- and high-semantic constraint sentences in English, Spanish, French, and German, manipulating the degree of semantic constraint to investigate its influence on the models' efficacy in bilingual processing. The target interlingual homographs comprise pairs across English-Spanish, English-French, and English-German language pairs in our dataset. Each sentence presents an incongruent condition, where the language of the sentence and the context differ from the language of the target word language and meaning. 
This introduces a cross-lingual ambiguity, where the meaning of the context and the target word differ, as presented in Example \ref{example:task3}, We want to analyse the model's ability to deal with this ambiguity. The model ideally should use the semantic cues present in the sentence and the orthographical similarities of the homograph pairs to understand the ambiguous sentence.
% This introduces an ambiguity in cross-lingual ambiguity, we want to analyse the model's ability to deal with this ambiguity, by leveraging the semantic cues present in the sentence and orthographical similarities that homograph pairs have\todo{modify this sentence}. 
Further details about how low and high-semantic constraint sentences are generated and curated are specified in Section \ref{sec:material}. Our task involves three paradigms.
\begin{enumerate}
    \item \textbf{Language identification.} We evaluate the model's ability to identify the language of a word within the code-mixed sentence. This is particularly challenging given that the word is orthographically similar to a homograph in the sentence's language.
    
    \item \textbf{Semantic understanding.} We assess how the model interprets the intended meaning of a word within its context under the congruent condition. Specifically, we focus on how the models disambiguate the ambiguity presented by the homograph.
    
    \item \textbf{Sentence comprehension.} We analyse whether the model can understand and make sense of the entire sentence, considering both semantic and syntactic cues.
\end{enumerate}

We do so by instructing the model using the instructions as depicted in Example C.

% \vspace{0.7mm}
% \newline
% \noindent
\begin{tcolorbox}[
colframe=black!50!black, 
colback=black!10, 
title=Example C: Instructions and example of sentences,
label=example:task3]
Given the following code-mixed sentence in English-Spanish, answer the following questions about the word enclosed in double quotes:
\begin{enumerate}
    \item What is the language of the enclosed word?
    \item What is the meaning of the enclosed word?
    \item Does the given sentence make sense in context with the meaning of the enclosed word?
\end{enumerate}

\dotfill

\textbf{Low-semantic constraint:} She bought a colorful \emph{``balón''} for her child.

\textbf{High-semantic constraint:} The \emph{``balón''} floated gently into the sky after it slipped from her grasp at the fair.

In both cases,  the language of the sentence (i.e. English) differs from the language in which the target homograph is presented (i.e. Spanish)

\end{tcolorbox}
We considered human evaluation to assess the model's response. \textcolor{black}{This constraint limited our evaluation to only the LLaMA-3.1 models. However, given the inherent ambiguity of the sentences, we found a manual evaluation procedure to be appropriate. Two annotators (a 24-year-old male and a 23-year-old female, both trained linguists) independently evaluated the LLM’s free-text responses. Their task was to interpret each response and map it onto predefined discrete labels, rather than to judge correctness directly or to provide independent answers.}

\textcolor{black}{Specifically, annotators were shown the LLM’s free-text generation for each item together with the two possible meanings of the interlingual homograph. For the word language-identification, they assigned the LLM’s response to one of the four languages in our corpus or an ``other" category; for the word meaning-identification question, they mapped the response to one of the two predefined meanings of the interlingual homograph or ``other". These labels were subsequently used to determine answer correctness.}

\textcolor{black}{In cases of disagreement between the two annotators, a third annotator (a 24-year-old male linguist) adjudicated. To support consistent interpretation, all annotators were provided with reference definitions for each homograph pair, enabling them to identify which sense the model intended to convey (\hyperref[SI:sub_section6]{Appendix E} contains further details).}\\

% However, given the inherent ambiguity of the sentences, we found the manual evaluation method appropriate. Two annotators (a 24-year-old male and a 23-year-old female, both trained linguists) were tasked with labelling the language that the LLM assigned to the word and the meaning it associated with that word. Specifically, the annotators were given the free-text generations of the LLM, along with the two possible meaning of the interlingual homograph and asked to either assign the word one of the four language or other label. In cases where the two annotators disagreed, a third annotator (a 24-year-old male linguist) was brought in to resolve the tie. To aid their judgments, all annotators were provided with the reference meanings of the homograph pairs, ensuring that they could accurately determine which sense of the word the model intended to convey (\hyperref[SI:sub_section6]{Appendix E} contains further details).\\

\noindent\textbf{Model's ability to identify a word's language.} We observe that LLaMA-3.1 can correctly identify the language of interlingual homographs with an accuracy of 72.98\% for English, 75.36\% for French, 68.57\% for German, and 77.14\% for Spanish. These results demonstrate the model's comparable performance across English, French, German, and Spanish, indicating a balanced capability in identifying the language of homographs. When the model does not assign the word with its own language label, it utilises the language of the context to the word as reflected in Figure~\hyperref[fig:main_t3_1]{3(A)}. This indicates the ability of the model to base its judgment on the surrounding context cues. We also note a few outliers in German homographs that the model identifies the word to be Dutch or Greek; these are labelled as neither in the figure.  \\

\noindent\textbf{Word meaning recognition.} Figure~\hyperref[fig:main_t3_1]{3(B)} shows the relation between the language of the interlingual homograph and the word meaning the model utilises in understanding the ambiguous sentence. For homographs in French, German, and Spanish, we note that the model utilises the English meaning in 69\%, 60\%, and 66\% of the cases, respectively. \textcolor{black}{Since these sentences are in English and semantically constrained by the English homograph}, this indicates the model's ability to align its understanding with the semantic cues in the context to understand the contextually relevant English meaning. However, English homographs do not show such trends -- 62\% of the words rely on the English meaning even though the context is contained in the corresponding non-English language. We also find the model to use neither of the homographs' meaning in $\sim8\%$ of the cases marked by `Neither' entry\\

\noindent\textbf{Model's word language identification ability and its meaning utility.} We also inspect the interplay between the model's ability to identify the language of the words and the meaning they utilise to disambiguate the sentence (c.f. Figure~\hyperref[fig:main_t3_1]{3(C)}). We note that 55\% of the homographs that are identified as Spanish, French or German are interpreted with English meaning. However, words assigned an English label are predominantly (75\% cases) interpreted with English meaning only. \\

\noindent\textbf{Model's correction ability. } Further, we analyse the model's correction ability in dealing with ambiguous sentences. Given the semantic constraint, the language of the context and the orthographical similarity of the homographs, the model should ideally correctly interpret the ambiguity by utilising the meaning of its counterparts. Hence, we draw the following scenarios of the model processing.

\begin{enumerate}
    \item \textbf{Correction Type-1:} This correction type focuses on the model's utility of word language identification to disambiguate. That is when the model identifies the word's language to be the same as that of the context and further utilises the correct semantically constrained meaning of the sentence.  
    \item \textbf{Correction Type-2:} This correction corresponds to the utility of context to correct the word. When the model identifies the homograph language to be dissimilar to that of the context, but utilises the meaning of the word within the bounds of the sentence.
    \item \textbf{Confusion:} \textcolor{black}{As shown in Figures~\hyperref[fig:main_t3_1]{3(A)} and ~\hyperref[fig:main_t3_1]{3(B)}, there are instances where the model assigns either a language or a word meaning that does not correspond to either of the two intended languages. We categorise such cases as Confusion. This label captures situations where the model’s prediction falls outside the valid set of labels, indicating that the model could not correctly ground the word in either language or meaning.
    \item \textbf{No Correction}: When the model identifies the word's language correctly and further utilises the word's meaning without considering the context.}
\end{enumerate}

\begin{figure*}[t!]
\centering
\includegraphics[width=\linewidth]{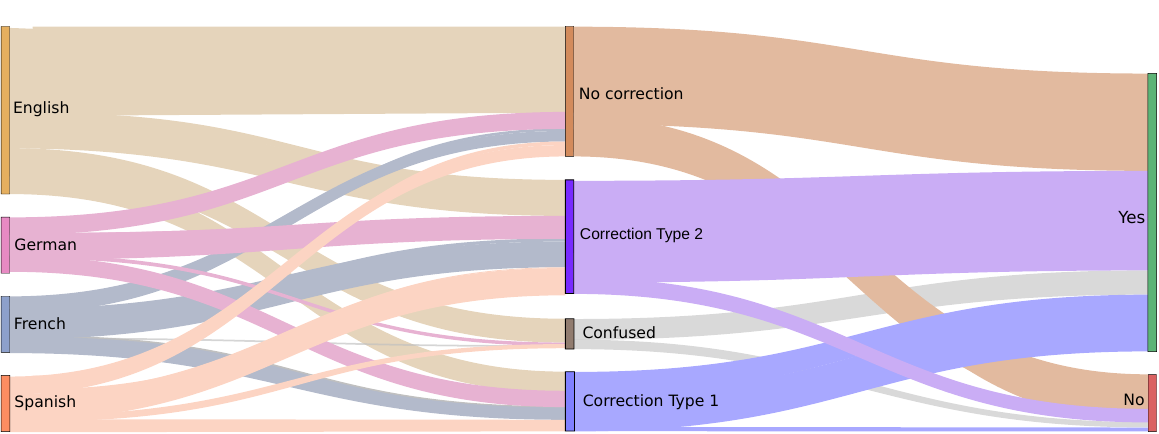}
\caption{\textbf{Model's correction ability:} The figure illustrates the model's correction and sentence comprehension abilities. We observe that the majority of non-English sentences undergo Correction Type 2, where the model uses context cues to correct the homograph. In contrast, English sentences undergo no correction, as the model relies on the word's meaning to understand the sentence, even though the condition is congruent. Furthermore, the model labels the sentence as semantically meaningful in most cases, regardless of whether it has been corrected.}
\label{fig:main_t3_2}
\end{figure*}
We note that the model successfully resolves the ambiguity of French, German, and Spanish homographs constrained in English sentences in 70\% cases (see Figure~\hyperref[fig:main_t3_2]{4} and \hyperref[SI:sub_section7]{Appendix F}), with Correction Type-2 being the most frequently employed method, used in 47.53\% of cases for these words. However, for English homographs, the model performs corrections in only 34\% cases (see \hyperref[SI:sub_section7]{Appendix F}), reflecting the multilingual model's inability to effectively leverage non-English sentences to infer the correct meaning and perform corrections. However, English homographs that do undergo correction (34\% of homographs) also utilise Correction Type-2 majorly. The utility of Correction Type-2 highlights the model's ability to contextualise concepts from the sentences to disambiguate non-English homographs and accurately interpret word meanings. 
Further, we note that high-semantic sentences undergo correction more than low-semantic sentences (c.f. Figure ~\ref{fig:hs_ls} in \hyperref[SI:sub_section8]{Appendix G}).

English homographs do not undergo any form of correction in 51\% of cases, utilising the meaning of the word without considering contextual cues (see Figure~\hyperref[fig:main_t3_2]{4} and \hyperref[SI:sub_section8]{Appendix G}). In contrast, French, German, and Spanish words remain uncorrected in 23.6\%, 28.2\% and 26.8\% cases, respectively. Therefore, non-corrected sentences are predominantly composed of English words, further reflecting the model's inability to reason within a non-English context. Furthermore, we note that most of the sentences in which the model gets confused correlate to English words (\hyperref[SI:sub_section8]{Appendix G} for more details). The sentences the model does not correct are marked as semantically meaningful by the LLM in 76\% of cases, while the sentences that undergo a correction are marked as meaningful in 92\% cases.
% \todo{
 % 1. Spseem to guide ecify percentage of words that didn't go through any correction. 
% 2. Specify how many non-corrected words beloend to English that were ultimately marked as semantically meaningful. Show a comparison.
% 3. Show that more Non-English words are corrected to fit the English context.
% }

\section{Discussion and Conclusion} In recent years, psycholinguistic research on lexical access and the mental representations of word-meaning associations has transformed to account for the universal, human capacity for multilingualism.  Instead of monolingual explanations of the cognitive processes that operate within the mental lexicon, we now have adaptive, dynamic models, such as the Bilingual Interactive Activation Plus (BIA+) Model~\cite{dijkstra2002architecture} that extend such explanations in the context of multilingualism and cross-lingual processing. With respect to the representation of cognates and interlingual homographs in the integrated bilingual lexicon, the BIA+ Model highlights the importance of orthographic, phonological, and semantic processing of the received input~\cite{jared2001bilinguals,10.1162/coli_a_00542,dijkstra2002architecture}. 

For orthographic processing alone, the model proposes that when a bilingual speaker is presented with an input letter string, several lexical orthographic candidates are simultaneously activated. This activation crucially depends on the level of orthographic similarity between the input string and each candidate, and the latter's resting level of activation (as determined by factors like frequency, recency of use, etc.)~\cite{dijkstra2002architecture,10.1162/coli_a_00342,Gervain2013}. This is relevant as we interpret the performance of multilingual LLMs on the three experimental tasks. The results obtained in our first task especially show that orthographic similarities and neighbourhood density crucially influence the models' ability to disambiguate different word types. When the models were presented with completely identical word pairs, their cognate status worked in favour of their high neighbourhood density and orthographic similarity. This suggests that the semantic overlap between cognates aids in faster word recognition when the orthographic similarity between them is incomplete, and even more so when they are completely identical. This is similar to the cognate facilitation effect, as explained by the BIA+ model, regarding bilingual speakers optimising available orthographic information to process cognates faster. Interestingly, our results also show how neighbourhood density can impede word recognition in multilingual LLMs, particularly in the case of interlingual homographs. Once again, this aligns with the BIA+ prediction that the simultaneous activation of orthographically similar candidates can be disadvantageous for word processing when they share distinct meanings across two languages. Indeed, as observed in our findings, LLMs are far better at processing partially identical homograph pairs as compared to identical ones.

However, we note that orthographic information only helps word processing to a certain extent. Our findings of the semantic judgment task show that the cognate facilitation effect does not persist and that LLMs are equally poor in storing and retrieving the meanings of cognates and homographs. This can be correlated with the lack of referential grounding in the LLMs training regime. The word-world view of the model is not correlated, leading to no correlation between orthological and semantic retrieval, which in humans are facilitated by each other. 

\textcolor{black}{Further, as explained by \citet{dijkstra2002architecture} in the architecture of the BIA+ Model, the input string’s language label exerts little influence in this initial identification process.} This underscores the non-selective nature of bilingual lexical access. BIA+ divides bilingual processing into a word identification system and a decision-making system. It is assumed that the decision-making system is influenced by nonlinguistic factors, such as task demands and extraneous variables. On the other hand, bilingual word identification is affected by linguistic context effects that arise from lexical, syntactic, or semantic sources (e.g., sentence context). When such linguistic information from different languages is made available, it influences the degree to which language-selective access is possible while trying to determine the meaning of interlingual homographs. Apart from this, the BIA+ model does not deem language labels or information about the language of the target word to play a major role in the early stages of bilingual word processing. This is because there can be many orthographic candidates feeding activation to a single language node, but the reverse activation from the node to these candidates is typically diluted. The delay in receiving such language information is apparent in the case of processing interlingual homographs, where there is far too much interference from non-target-language candidates competing for selection. The utility of Correction Type-2 in multilingual LLMs mimics this lack of influence of language label in word processing in the BIA+ model. As in Correction Type-2, while the word is identified as dissimilar from the context, the model utilises the context to retrieve a semantically correct meaning of the word. This correction, though preferred over correction type-1, is not predominantly used in word processing in our third setup. Furthermore, we notice a stark difference in how English and non-English homographs are processed. Hence, the LLM, unlike the BIA+ model, is not completely independent from the utility of language labels in word processing, especially if the word is in English. More alignment across all languages is essential for better alignment of LLMs and human bilingual processing.

\newpage
\appendix

\appendixsection{Significance Study}
\label{SI:sub_section1}

To compare the statistical significance of the multilingual LLM's ability to disambiguate word pairs and make semantic judgments, we compare it against a random baseline. The random model assumes a uniform probability distribution over the possible label space (i.e., \textit{True} and \textit{False} for the word disambiguation task, and `1' or `2' for the semantic judgment task). The predictions of the models are obtained by taking the mode of the five seed runs. We utilise Welch's t-test~\cite{welch1947generalization} and calculate the $p$-value to evaluate the null hypothesis that there is no significant difference between LLMs and the random model. Our results, as presented in Table~\ref{tabel:sig_t1} for word pair disambiguation and Table~\ref{tabel:sig_t2} for semantic judgment, show that the $p$-value for all models across both tasks is $<0.05$. This indicates a statistically significant difference between the models and the random baseline.

% \begin{table*}[2][h]
\begin{table*}[h]
\centering
\resizebox{0.8\textwidth}{!}{ 
\begin{tabular}{c|ccccc}
\hline
\backslashbox{Lang}{Model} & BLOOMZ& LLaMA2 & LLaMA3 & LLaMA3.1 & MISTRAL \\ \hline
English-German  & \cellcolor{green!40}$2.34\times 10^{-10}$ 
                & \cellcolor{green!40}$8.14\times 10^{-9}$ 
                & \cellcolor{green!40}$9.41\times 10^{-8}$ 
                & \cellcolor{green!40}$4.36\times 10^{-7}$ 
                & \cellcolor{green!40}$6.34\times 10^{-8}$ \\

English-Spanish & \cellcolor{green!40}$1.40\times 10^{-4}$ 
                & \cellcolor{green!40}$6.10\times 10^{-5}$ 
                & \cellcolor{green!40}$6.10\times 10^{-5}$ 
                & \cellcolor{green!40}$1.10\times 10^{-4}$ 
                & \cellcolor{green!40}$1.20\times 10^{-3}$ \\

English-French  & \cellcolor{green!40}$1.33\times 10^{-7}$ 
                & \cellcolor{green!40}$2.88\times 10^{-7}$ 
                & \cellcolor{green!40}$9.05\times 10^{-8}$ 
                & \cellcolor{green!40}$1.25\times 10^{-6}$ 
                & \cellcolor{green!40}$6.08\times 10^{-7}$ \\ 
\hline
\end{tabular}
}
\caption{Significant score for word pair disambiguation task for the three language pairs -- English-German, English-Spanish, and English-French. Cells are colour-coded according to statistical significance (p < 0.05): green indicates significant values, and red indicates non-significant values. In this table, all values are statistically significant.}
\vspace{-5mm}
\label{tabel:sig_t1}
% \end{table*}
\end{table*}

% \begin{SCtable*}[2][h]
\begin{table*}[h]
\centering
\resizebox{0.8\textwidth}{!}{ 
\begin{tabular}{c|ccccc}
\hline
\backslashbox{Lang}{Model} & BLOOMZ& LLaMA2 & LLaMA3 & LLaMA3.1 & MISTRAL \\ \hline
German  & \cellcolor{green!40}$2.00\times 10^{-3}$  & \cellcolor{green!40}$3.80\times 10^{-2}$ & \cellcolor{green!40}$3.00\times 10^{-2}$ & \cellcolor{green!40}$4.60\times 10^{-2}$ & \cellcolor{green!40}$3.20\times 10^{-2}$ \\ 
English & \cellcolor{green!40}$2.05\times 10^{-14}$ & \cellcolor{green!40}$1.35\times 10^{-13}$ & \cellcolor{green!40}$1.52\times 10^{-12}$ & \cellcolor{green!40}$4.06\times 10^{-9}$  & \cellcolor{green!40}$1.41\times 10^{-16}$ \\ 
Spanish & \cellcolor{green!40}$8.30\times 10^{-5}$  & \cellcolor{green!40}$1.00\times 10^{-4}$  & \cellcolor{green!40}$4.00\times 10^{-5}$  & \cellcolor{green!40}$1.00\times 10^{-4}$  & \cellcolor{green!40}$3.00\times 10^{-4}$ \\ 
French  & \cellcolor{green!40}$1.00\times 10^{-3}$  & \cellcolor{green!40}$3.00\times 10^{-3}$ & \cellcolor{green!40}$2.20\times 10^{-2}$ & \cellcolor{green!40}$1.00\times 10^{-3}$  & \cellcolor{green!40}$1.00\times 10^{-2}$ \\ \hline

\end{tabular}
}
\caption{Significant scores for the semantic judgment task. Cells are colour-coded according to statistical significance (p < 0.05): green indicates significant values, and red indicates non-significant values. In this table, all values are statistically significant.}
\vspace{-5mm}
\label{tabel:sig_t2}
\end{table*}
% \end{SCtable*}

\appendixsection{Performance of Shots and Models in Word Disambiguation}
\label{SI:sub_section2}

\begin{figure*}[t!]
\centering
\includegraphics[width=\textwidth]{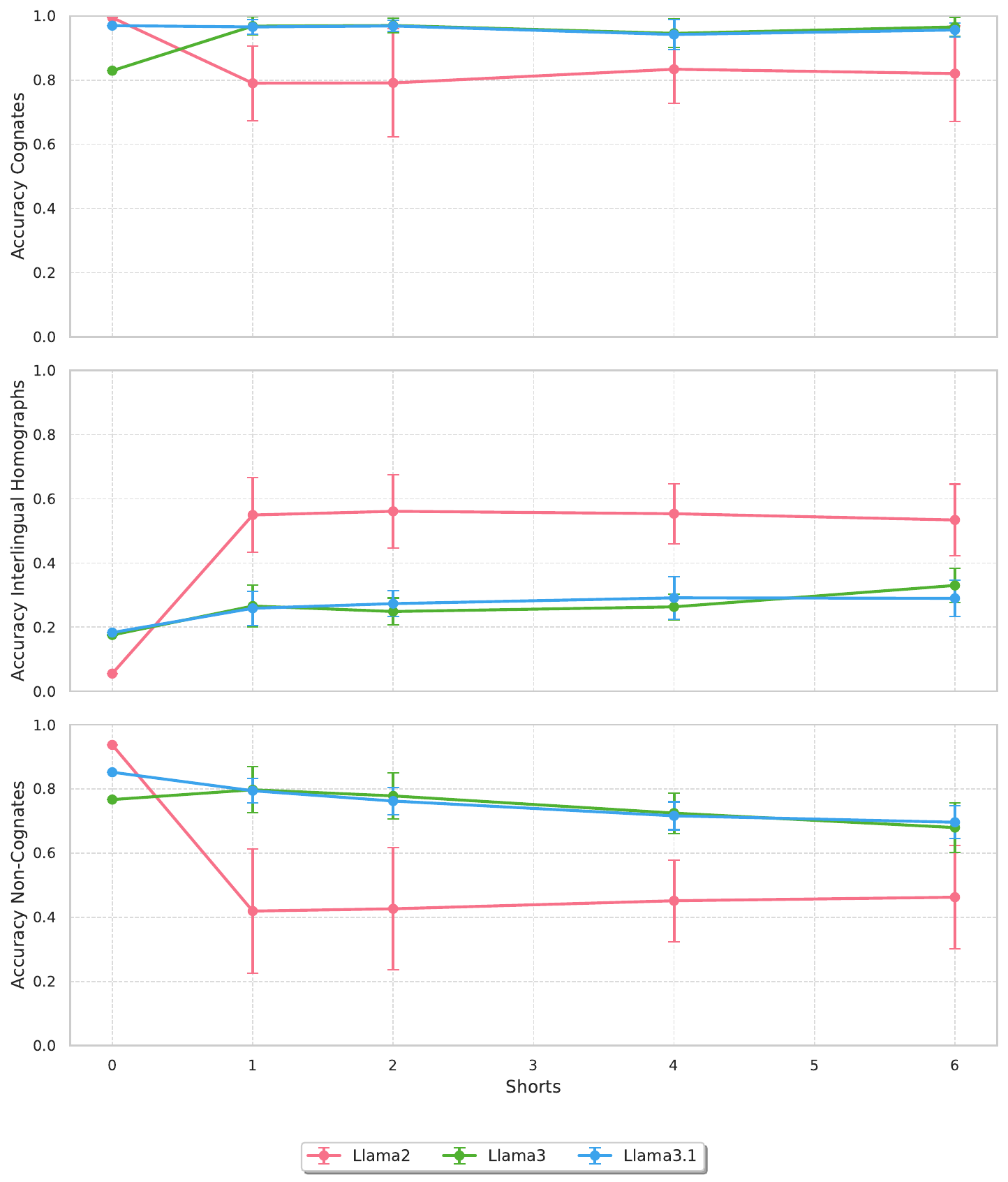}
\caption{\textbf{Variation in performance across shots.} \textbf{(A) Cognates:} We note the performance improves with the number of shots and stabilises after two shots. LLaMA-3.1 consistently outperforms the other models, while LLaMA-2 shows more variation. \textbf{(B) Non-cognates:} The performance of models decreases with the number of shots; however, for LLaAMA-2, the performance increases after shot two. \textbf{(C) Interlingual Homographs:} Both LLaMA-3 and LLaMA-3.1 show steady improvement with additional shots, whereas LLaMA-2 exhibits substantial variability but achieves its highest performance by four shots.}
\label{fig:shots}
\end{figure*}

The performance of LLMs improves as the number of demonstrations provided in the prompt increases~\cite{brown2020language}. Hence, to find an ideal number of demonstrations, we run an ablation experiment, varying the number of demonstrations in the word pair disambiguation task. Figure~\ref{fig:shots} shows the performance of cognates, non-cognates, and interlingual homographs on five different shots, where each shot represents the number of \textit{True} and \textit{False} label examples used in the demonstration. We observe that the performance of cognates and interlingual homographs improves with the number of shots, stabilising largely after two shots. However, for non-cognates, we observe a decline in performance after four shots. We also note that the variation in performance across all word types decreases as the number of shots increases. Based on these observations, we prompt the model with four demonstrations
% \todo{what is it?} 
in word disambiguation and semantic judgment tasks.

\appendixsection{Word Pair Disambiguation}
\label{SI:sub_section3}

\begin{figure*}[t!]
\centering
\includegraphics[width=\textwidth]{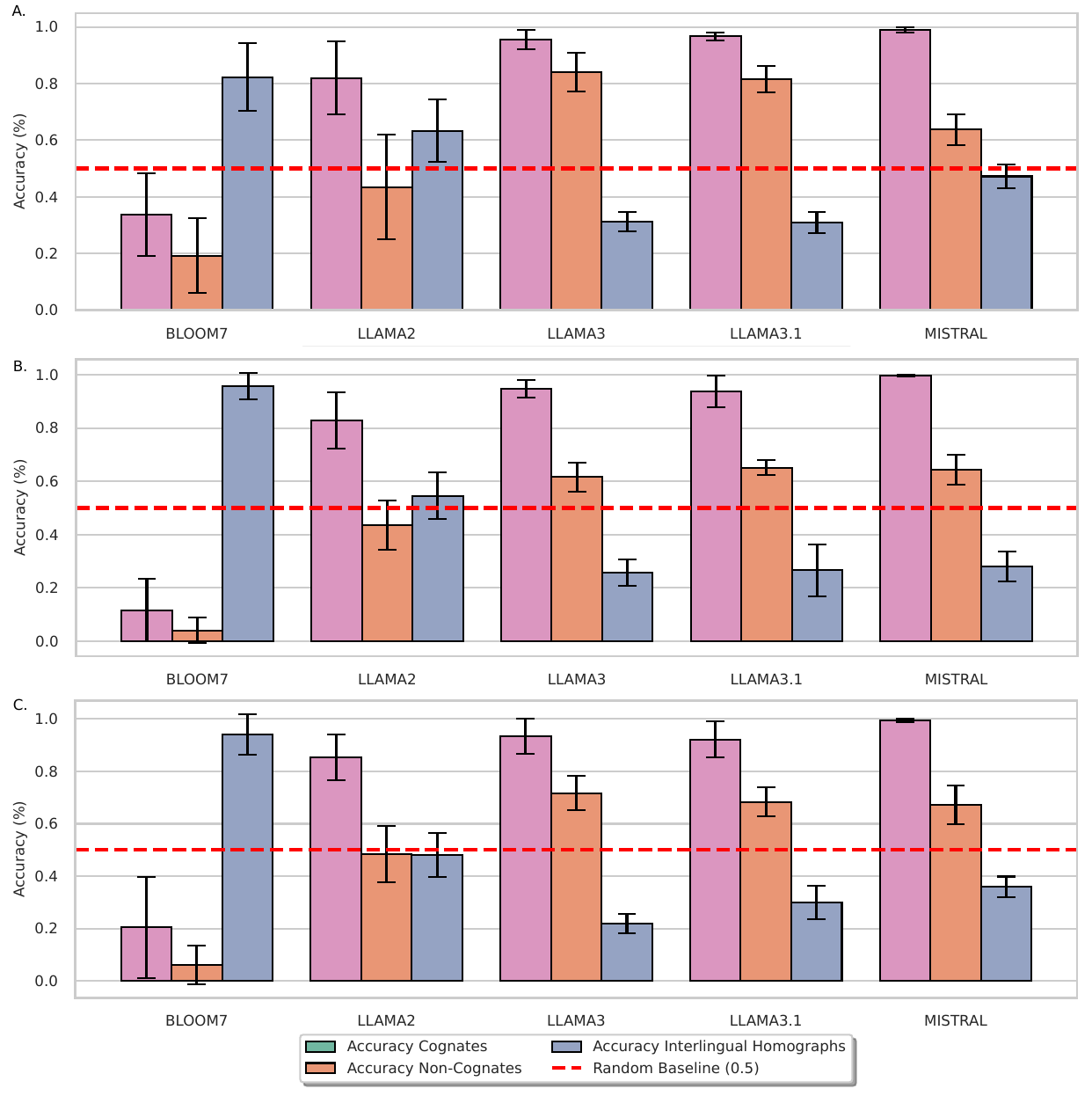}
\caption{\textbf{Word pair disambiguation accuracy.} The figure shows the performance of five multilingual LLMs, on five seed values, in identifying word pairs to have the same meaning or not on three language pairs -- \textbf{(A) English-German, (B) English-Spanish and (C) English-French}. All the models seem to perform better on cognate than on Non-cognates pairs, highlighting the role of cognate facilitation in LLMs. All models except BLOOMZ seem to perform poorly in disambiguating phonograph pairs as compared to cognates, further showing their utilisation of orthographical signals in the disambiguation task.}
\label{fig:word-dis_lang}
\end{figure*}

Figure~\ref{fig:word-dis_lang} shows the performance of five models: BLOOMZ, LLaMA-2, LLaMA-3, LLaMA-3.1, and Mistral on the three language pairs, English-German, English-Spanish, and English-French, of our datasets. We observe the role of cognate facilitation in all models in all language pairs, with overall performance on cognates being 78\% and on non-cognates being 52\%.

\appendixsection{Semantic Judgment and Word Pair Disambiguation Correlation}
\label{SI:sub_section4}

% \begin{SCtable*}[2][h]
\begin{table*}[h]
\centering
\resizebox{0.9\textwidth}{!}{ 
\begin{tabular}{l|cc|cc|cc}
\hline
\backslashbox{Entropy Type}{Model} & LLaMA3.1 & MISTRAL& LLaMA3.1 & MISTRAL& LLaMA3.1 & MISTRAL \\ \hline
\multicolumn{1}{c}{\cellcolor[HTML]{EEEEEE}} &
\multicolumn{2}{c}{\cellcolor[HTML]{EEEEEE}English-Spanish}&
\multicolumn{2}{c}{\cellcolor[HTML]{EEEEEE}English-German}&
\multicolumn{2}{c}{\cellcolor[HTML]{EEEEEE}English-French}\\ \hline

Entropy of word dis (H(word dis))&0.6972&0.7519&0.6471&0.8564&0.7328&0.7612\\
Entropy of sem jud (H(sem jud))&0.5917&0.5652&0.5719&0.6292&0.6471&0.5157\\
Conditional Entropy H(sem jud | word dis)&0.5905&0.5647&0.5716&0.6284&0.6471&0.5157\\
Conditional Entropy H(word dis | sem jud)&0.6960&0.7515&0.6468&0.8555&0.7328&0.7612\\\hline

\end{tabular}
}
\caption{Entropy and conditional entropy of word pair disambiguation (word dis)  and semantic judgment (sem jud).}
\vspace{-5mm}
\label{tab:entropy}
\end{table*}
% \end{SCtable*}

We utilise conditional entropy to measure the correlation between the semantic judgment task and the word pair disambiguation task~\cite{ash2012information}. Conditional entropy, $H(Y|X)$, measures the uncertainty in a random variable $Y$ given the value of the random variable $X$ is known. It is defined as: 
$$H(Y|X)=\sum_{x \in X} P(x) H(Y|X = x) \text{ where, } H(Y|X=x) = -\sum_{y \in Y} P(y|x) \log P(y|x)$$
If the conditional entropy, $H(Y|X)$, is similar to the entropy of $Y$, $H(Y)$, it indicates that the two variables are independent. We define $H(Y)$ as:

$$H(Y) = - \sum_{i} p(y_i) \log(p(y_i))$$
Table~\ref{tab:entropy} shows the conditional entropy of the tasks. We observe that $H(\text{Semantic Judgment} \mid \text{Word Pair Disambiguation}) \sim H(\text{Semantic Judgment})$  and $H(\text{Word Pair Disambiguation} \mid \text{Semantic Judgment}) \sim H(\text{Word Pair Disambiguation})$,  indicating the independence of the two variables.

\appendixsection{Human Evaluation on LLM's Responses}
\label{SI:sub_section6}

Given the open-ended nature of the sentence disambiguation task, we employed two annotators, one male and one female, aged 24 and 23, respectively, to evaluate the responses of LLaMA-3.1. Since the responses were in English, the annotators were required to demonstrate high English proficiency. Each annotator was required to have achieved a minimum score of 8 on the IELTS or 100 on the TOEFL in the past year to ensure their competence in assessing the model's responses. The annotators were provided with the model's response along with the meaning of the interlingual homograph pair so that they could determine which meaning from the pair was conveyed by the model. We do not directly provide the interlingual homograph or the prompt of the LLM to the annotators. 

To assess the agreement between the two annotators, we used the Kappa score. The Kappa score is a statistical measure that quantifies the level of agreement between two annotations. It ranges from -1 to 1, formalised as:  
\[
\kappa = \frac{P_o - P_e}{1 - P_e}, 
\]
Where $P_o$ and $P_e$ characterise the observed and expected agreement, respectively.
The Kappa score of the annotators for English, French, German and Spanish homographs was 0.94, 0.96, 0.74 and 0.66, respectively, showing substantial agreement between the annotators. The responses in which the annotator disagreed were passed to a third annotator, a male, aged 24, to break the tie.

\appendixsection{Correction type distribution}
\label{SI:sub_section7}

\begin{figure*}[!t]
\centering
\includegraphics[width=0.7\textwidth]{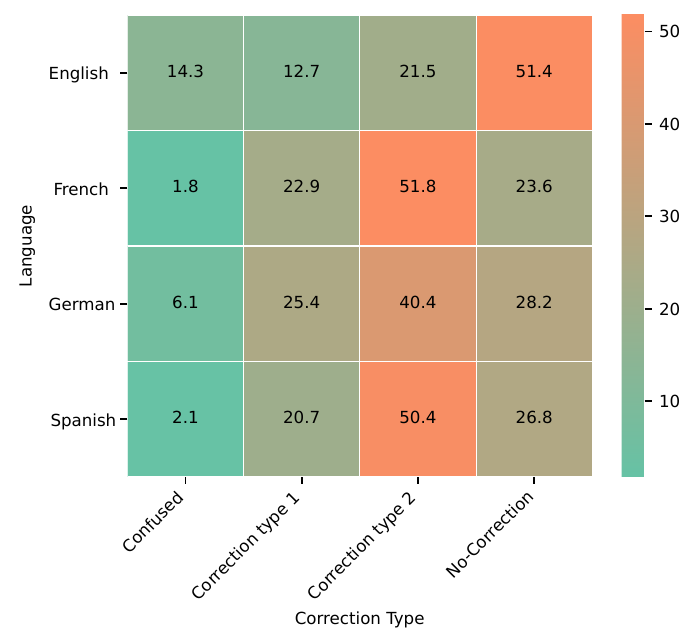}
\caption{\textbf{Correction type distribution.} The figure shows how homographs are processed by LLaMA-3.1 in the sentence disambiguation task.}
\label{fig:correction_heatmap}
\end{figure*}

Figure~\ref{fig:correction_heatmap} shows the distribution of correction types utilised by LLaMA3.1 in the sentence disambiguation task, for each language in our dataset: English, French, German, and Spanish. We observe that for 51\% of the English homographs, the model does not undergo any form of correction. In contrast, for non-English languages, the model utilises Correction Type 2 in 47.53\% of cases. This correction type corresponds to recognising the language identity of the homograph but utilising the English sentence's context to infer the homograph's meaning to disambiguate the sentence. This highlights the inability of the models to deal with non-English sentences~\cite{zhang2023don}.

\appendixsection{Homograph's Distribution Across Correction Type}
\label{SI:sub_section8}

\begin{figure*}[!ht]
\centering
\includegraphics[width=\textwidth]{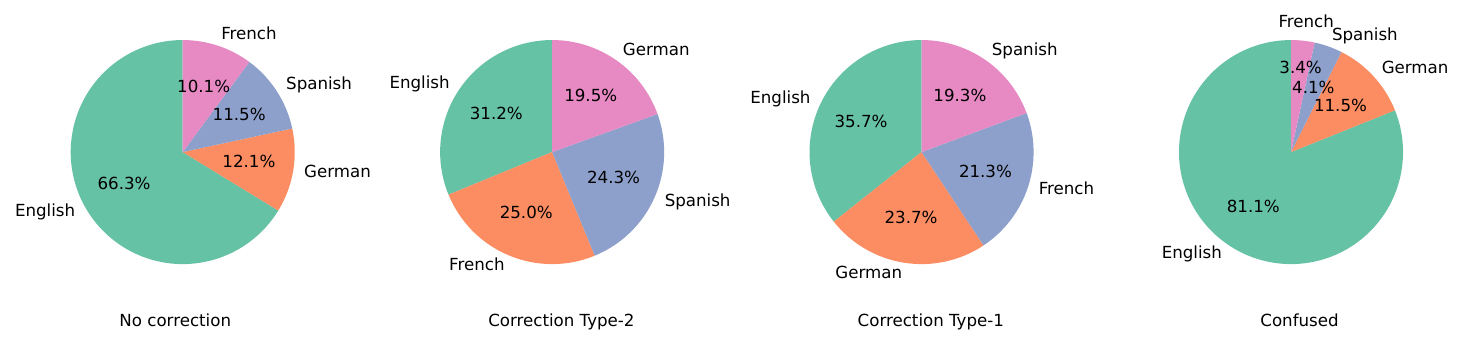}
\caption{\textbf{Language distribution across the four correction types.} The figure shows the homograph language distribution, in the four correction types: No-correction, Correction Type-1, Correction Type-2, and Confused.}
\label{fig:pie_distribution_correction}
\end{figure*}

Figure~\ref{fig:pie_distribution_correction} analyses the language distribution of homographs in each of the correction types. Notably, 66.3\% of the No-correction category corresponds to English homographs, whereas, for non-English languages -- French, German, and Spanish -- correspond to 10.1\%, 11.5\%, and 12\%, respectively. Furthermore, we note that English words make up the majority (81.1\%) of `Confused' types.

\begin{figure*}[!ht]
\centering
\includegraphics[width=\textwidth]{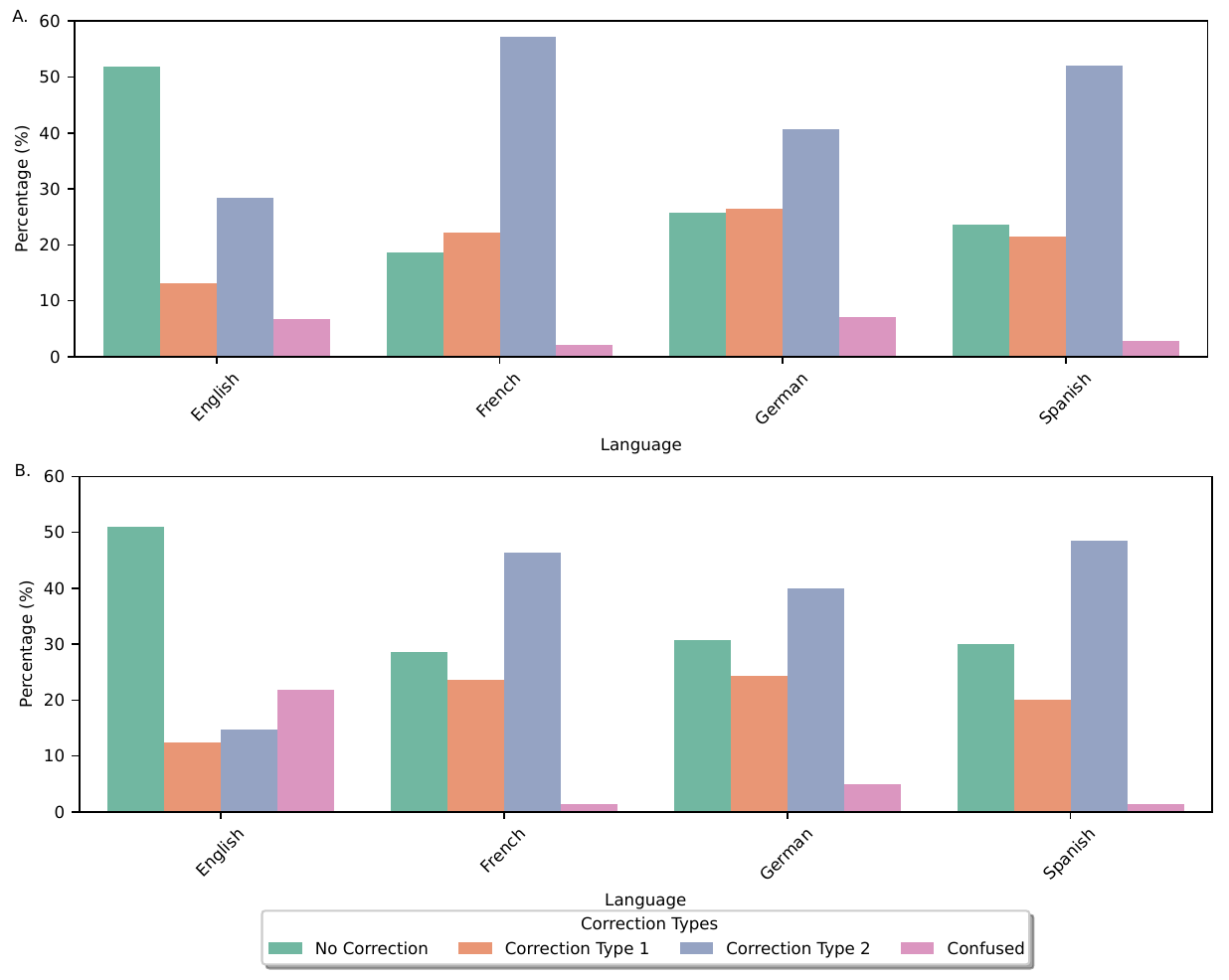}
\caption{\textbf{Distribution of correction type across the four languages.} The figure shows the distribution of corrections in the four languages: English, French, German, and Spanish, in A.) high-semantic and B.) low-semantic constraints.}
\label{fig:hs_ls}
\end{figure*}

\textcolor{black}{Figure~\ref{fig:hs_ls} also note that high-semantic constraint undergoes more correction than low-semantic sentences.}

% \begin{figure*}[h]
% \centering
% \includegraphics[width=0.5\textwidth]{fig/correction_heatmap.pdf}
% \caption{\textbf{Correction type distribution.} The figure shows the performance of LLaMA3.1, on the sentence disambiguation task, in the four languages covered in our dataset: English, French, German, and Spanish. Notably, for 51\% of the English homographs, the model doesn't undergo any form of correction. Whereas, for non-English languages - French, German, and Spanish - the model utilities Correction type 2 in 47.53\% of cases, which corresponds to understanding the language identity of the language but utilising the sentence's context to infer its meaning to disambiguate the sentence. Furthermore, we note that English words are confused more compared to other languages.}
% \label{fig:word-dis_lang}
% \end{figure*}

\if 0
\begin{acknowledgments}
This is an example of acknowledgement. This is a sample paper presented just for the coding of different elements of a document in \LaTeX\ using this package file. We thank all readers.
\end{acknowledgments}
\fi 

%\newpage
%If you would like to use \starttwocolumn below for references: Unfortunately, there is a known bug in the existing \starttwocolumn command when used without \newpage above it. If you find issues such as whole pages disappearing, add \newpage above \starttwocolumn as a temporary fix. We will provide a permanent solution in a future update.
%\starttwocolumn

\bibliographystyle{compling}
\bibliography{COLI_template}

\end{document}